\documentclass[10pt,journal,compsoc]{IEEEtran}

\usepackage[colorlinks,anchorcolor=blue]{hyperref}
\usepackage{booktabs}       
\usepackage{xcolor}         
\usepackage{algorithm,algorithmic}
\usepackage{cite}

\usepackage{colortbl}
\usepackage{amsmath}
\usepackage{amssymb}
\usepackage{ntheorem}
\usepackage{enumitem}
\usepackage{graphicx}
\usepackage{ragged2e}  
\usepackage[caption=false,font=normalsize,labelfont=sf,textfont=sf,subrefformat=parens,labelformat=parens]{subfig}
\usepackage{multirow}
\usepackage{makecell}
\graphicspath{{figures/}}
\newcommand{\MC}{\mathcal}
\newcommand{\MBB}{\mathbb}
\newcommand{\MBF}{\mathbf}
\newcommand{\etal}{et al.}
\newcommand{\wrt}{w.r.t.}
\newcommand{\JS}{\mathrm{JS}}
\newcommand{\KL}{\mathrm{KL}}
\newcommand{\TV}{\mathrm{TV}}
\newcommand{\mar}{\mathrm{mar}}
\newcommand{\con}{\mathrm{con}}
\newcommand{\supp}{\mathrm{supp}}

\newtheorem{definition}{Definition}
\newtheorem{theorem}{Theorem}

\newtheorem{proposition}{Proposition}

\newtheorem{remark}{Remark}
\newcommand*\diff{\mathop{}\!\mathrm{d}}
\definecolor{fig_blue}{RGB}{31,119,180}
\definecolor{fig_orange}{RGB}{255,127,14}
\usepackage{thmtools}

\begin{document}

\title{When Invariant Representation Learning \\ Meets Label Shift: \\ Insufficiency and Theoretical Insights}

\author{You-Wei Luo and Chuan-Xian~Ren
\IEEEcompsocitemizethanks{\IEEEcompsocthanksitem Y.W. Luo and C.X. Ren are with the School of Mathematics, Sun Yat-Sen University, Guangzhou 510275, China. C.X. Ren is the corresponding author (email: rchuanx@mail.sysu.edu.cn).\protect\\
\IEEEcompsocthanksitem This work was supported in part by National Natural Science Foundation of China (Grant No. 61976229), in part by Guangdong Basic and Applied Basic Research Foundation (2023B1515020004), in part by Science and Technology Program of Guangzhou (2024A04J6413), in part by Guangdong Province Key Laboratory of Computational Science at the Sun Yat-sen University (2020B1212060032), and in part by the Fundamental Research Funds for the Central Universities, Sun Yat-sen University.}
}

\markboth{IEEE TRANSACTIONS ON PATTERN ANALYSIS AND MACHINE INTELLIGENCE}%
{Shell \MakeLowercase{\textit{\etal}}: Bare Demo of IEEEtran.cls for Computer Society Journals}


\IEEEtitleabstractindextext{
\begin{abstract}
\justifying
As a crucial step toward real-world learning scenarios with changing environments, \textit{dataset shift} theory and \textit{invariant representation learning} algorithm have been extensively studied to relax the identical distribution assumption in classical learning setting. Among the different assumptions on the essential of shifting distributions, generalized label shift (GLS) is the latest developed one which shows great potential to deal with the complex factors within the shift. In this paper, we aim to explore the limitations of current dataset shift theory and algorithm, and further provide new insights by presenting a comprehensive understanding of GLS. From theoretical aspect, two informative generalization bounds are derived, and the GLS learner are proved to be sufficiently close to optimal target model from the Bayesian perspective. The main results show the insufficiency of invariant representation learning, and prove the sufficiency and necessity of GLS correction for generalization, which provide theoretical supports and innovations for exploring generalizable model under dataset shift. From methodological aspect, we provide a unified view of existing shift correction frameworks, and propose a kernel embedding-based correction algorithm (KECA) to minimize the generalization error and achieve successful knowledge transfer. Both theoretical results and extensive experiment evaluations demonstrate the \textit{sufficiency and necessity} of GLS correction for addressing dataset shift and the superiority of proposed algorithm.
\end{abstract}

}

{\maketitle}

\IEEEdisplaynontitleabstractindextext
\IEEEpeerreviewmaketitle

\IEEEraisesectionheading{\section{Introduction}\label{sec:introduction}}

\IEEEPARstart{A}{iming} to relax the identical distribution assumption in standard learning scenario, \textit{dataset shift}, also known as distribution shift, has received increasing attention in machine learning, computer vision and statistics communities~\cite{quinonero2008dataset,cai2021transfer}. In dataset shift scenario, the primary goal is to learn an invariant model for the potentially changing real-world environments, which is closely connected with \textit{domain adaptation} (DA) problem \cite{zhang2013domain,ganin2016domain,long2018transferable,courty2016optimal,zhang2019optimal,luo2022unsupervised,ren2022buresnet}. Specifically, the model trained on source domain (distribution) $P$ with sufficient knowledge (e.g., annotations) is supposed to be unbiased on a related but different target domain (distribution) $Q$ with less or no prior knowledge, i.e., semi-supervised and unsupervised transfer. To explore mathematical understanding and effective learning framework for such a generalization problem, considerable efforts have been made for the advancements of dataset shift theory~\cite{ben2010theory,zhao2019learning,redko2019advances} and DA algorithm~\cite{ganin2016domain,long2018transferable,combes2020domain,kirchmeyer2022mapping}, which also show great potential to deal with the transfer/generalization problems in real-world problems, e.g., computer vision~\cite{patel2015visual}, natural language processing~\cite{miller2020effect}, medical diagnosis~\cite{xu2021cross}, data privacy~\cite{xia2021adaptive,liu2021source}.

To characterize the essential of shifting distributions, several assumptions have been made based on different factorizations of distributions. Most of the works formulate dataset shift as covariate shift~\cite{shimodaira2000improving,sugiyama2007covariate} or conditional shift~\cite{zhang2013domain,gong2016domain,luo2021conditional,ren2022buresnet}, where the distributions over covariate $X$ or $X$ given label $Y$ are shifting, i.e., $P_X\neq Q_X$ or $P_{X|Y} \neq Q_{X|Y}$, respectively. These assumptions are closely related to an important and popular framework called \textit{invariant representation learning}~\cite{ganin2016domain,long2018transferable,zhao2019learning,stojanov2021domain,zhao2022fundamental}, where the shifting distributions are supposed to be aligned in a latent space with a mapping $g: X\mapsto Z$. Another fruitful assumption is label shift~\cite{scholkopf2012causal,lipton2018detecting,garg2020unified,gu2021adversarial}. Label shift is previously observed in medical diagnosis and time series data, and recently studied in vision and recognition communities~\cite{zhang2013domain,gong2016domain,combes2020domain,rakotomamonjy2021optimal,kirchmeyer2022mapping}. It assumes that the label distributions (e.g., $P_Y$) are changing across domains while the concepts are invariant. Recently, some new insights into dataset shift are provided from the perspectives of information theory~\cite{zhao2019learning,federici2021information,li2021learning}, which explicitly imply that the invariant representation learning and label shift correction are closely related and non-negligible. By considering the interaction of invariant representation learning and label shift correction, theoretical results~\cite{zhang2013domain,combes2020domain,kirchmeyer2022mapping} are derived under a more practical setting called generalized label shift (GLS). Several recent advancements~\cite{ren2018generalized,yan2017mind,long2018conditional,liu2021adversarial,rakotomamonjy2021optimal} in transfer/generalization also show the persistent concerns on the precise characterization and effective modeling for the GLS problem. 

Though GLS has shown great potential to deal with the complex factors in real-world shift scenarios, a systematic study and unified view of \textit{invariant representation learning} and \textit{dataset shift theory} are still lacking in current research, which is indeed necessary to show the superiority of GLS correction. In this paper, we aim to provide new insights from both theoretical and methodological aspects. \textit{Theoretical aspects}: motivated by the impossibility result \cite{zhao2019learning} for the transformation $g$ with marginal invariant property, we further analyze the limitations of invariant representation learning {\wrt} different types of transformation and provide comprehensive understanding of GLS problem. The key is to theoretically incorporate invariant transformation and label shift correction into the learning model under dataset shift. Compared with recent theoretical advances on the generalization upper bound of GLS~\cite{combes2020domain,kirchmeyer2022mapping}, our results not only ensures a more explicit upper bound, which directly characterizes the essential of GLS, but also prove the necessity of GLS correction by deducing an information-theoretical lower bound. Besides, our results show the insufficiency of invariant representation learning and prove that invariant representation model is error-prone under label shift, which also reveal the importance of understanding GLS. \textit{Methodological aspects}: we propose a general learning method for GLS correction based on the theoretical results, which provides a unified view of existing distribution shift frameworks. A kernel embedding-based correction algorithm (KECA) with explicit discrepancy optimization is proposed, which show superior performance in empirical validations. Overall, our contributions can be summarized as follows.
\begin{itemize}
  \item The limitations of invariant representation learning are further analyzed by extending the impossibility result to different invariant learning frameworks, and showing the essential conflict between them.
  \item A comprehensive understanding of GLS is presented by rigorously defining the learning model as a triplet. Further, the \textit{sufficiency and necessity} of GLS correction for successful transfer/generalization is proved by the derived generalization bounds.
  \item Intuitive interpretations are provided from the perspectives of Bayes error rate which ensures that the error of optimal GLS correction model can be sufficiently close to the lowest error rate on target domain.
  \item A general learning principle with optimization algorithm for GLS correction is proposed, and its connections with existing dataset shift correction frameworks are analyzed. The validity of derived theoretical results and proposed algorithm are verified by extensive experiments and analysis.
\end{itemize}

The rest of this paper is organized as follows. The preliminary results on dataset shift theory and invariant representations algorithms are reviewed in Sec.~\ref{sec:preliminary}. In Sec.~\ref{sec:main_results}, the main results on limitations of invariant representation learning and theoretical insights into GLS correction are presented. A general framework for GLS correction and the kernel-based learning algorithm are proposed in Sec.~\ref{sec:Learn_Prin_Alg}. The theory and proposed algorithm are validated by extensive experiments and analysis in Sec.~\ref{sec:experiment}. Finally, the conclusion and future direction are presented in Sec~\ref{sec:conclusions}.

\section{Preliminary}
\label{sec:preliminary}

In this section, we briefly review the recent results on dataset shift theory in Sec.~\ref{subsec:preliminary_theory}, and discuss the related invariant representation learning algorithms in Sec.~\ref{subsec:preliminary_theory}.

\textbf{Notations.}
We characterize the domains from the perspectives of statistics. The source and target domains are denoted by distributions $P$ and $Q$ over input (e.g., image) $X$ and output (e.g., label) $Y$ which take values from spaces $\MC{X}$ and $\MC{Y}$, respectively. For simplicity, let the lowercase letters $p$ and $q$ denote the probability density functions (PDFs). The subscripts denote the corresponding variables, e.g., $P_{XY}$ represents joint distribution. Let $d_{\KL}(\cdot \| \cdot)$, $d_{\JS}(\cdot,\cdot)$ and $d_{\TV}(\cdot,\cdot)$ denote the Kullback-Leibler (KL) divergence, Jensen-Shannon (JS) divergence and total variation (TV) distance between probability distributions. For any constant $c \in (0,1)$, the generalized JS divergence~\cite{lin1991divergence,nielsen2019jensen,cai2022distances} for mixture distributions is defined as $d_{\JS,c}(P,Q)= (1-c)d_{\KL}(P\| \mu) + c d_{\KL}(Q\| \mu)$, where $\mu = (1-c)P + c Q$. The detailed definitions are provided in \textit{supplementary material}.

\subsection{Dataset Shift Theory}
\label{subsec:preliminary_theory}

\textbf{Standard Learning Scenario.}
In classical learning scenario, the distributions $P$ and $Q$, which usually represent training and test distributions, are assumed to be the same. Given a hypothesis class $\MC{H}$, the basic goal is to learn a hypothesis $h\in \MC{H}$ on the accessible source distribution $P$. Let $\ell(\cdot,\cdot): \MC{Y}\times\MC{Y} \rightarrow \MBB{R}_{+}$ be the loss function, learning methods aim to minimize the \textit{true risk} as~\cite{shalev2014understanding}
\begin{equation}
  \label{eq:true_risk}
  \mathop{\arg \min}_{h\in\MC{H}} ~ \varepsilon_P(h) = \MBB{E}_{P_{XY}} \left[ \ell(h(X),Y) \right].
\end{equation}
Since $P=Q$, the expected true risk on $P$ is equivalent to $Q$, i.e., $\varepsilon_P(h)= \varepsilon_Q(h) $. 

\textbf{Dataset Shift Scenario.} 
Unfortunately, the identical distribution assumption generally does not hold for the changing real-world environments~\cite{scholkopf2012causal}. Such scenarios are called \textit{distribution/dataset shift}~\cite{quinonero2008dataset,redko2019advances}, which lead to a non-zero discrepancy between domains and the biased risk estimations $\varepsilon_P(h)$ for $Q$. To quantify the expected distance between $\varepsilon_P$ and $\varepsilon_Q$, theoretical results are derived by considering different types of shifts, e.g., covariate shift~\cite{ben2010theory,zhao2019learning,zhang2019bridging}, label shift~\cite{garg2020unified}, conditional shift~\cite{federici2021information} and GLS~\cite{combes2020domain,kirchmeyer2022mapping}.

For covariate shift, it usually considers a deterministic form of label as $Y=f(X)$, where $f$ is the labeling rule. Then the true risk can be rewritten as the distance between hypothesis and labeling rule, i.e., $\varepsilon_P(h,f_P) = \MBB{E}_{P_{XY}} [ \ell(h(X),f_P(X)) ]=\MBB{E}_{P_{X}} [ \ell(h,f_P) ]$. Several results for binary classification~\cite{ben2010theory,cortes2019adaptation,zhao2019learning} are derived based on the $\MC{H} \Delta \MC{H}$-distance and its variants, and further extended to multiclass classification by introducing the margin disparity discrepancy~\cite{zhang2019bridging}. These results decompose the generalization error as discrepancies on marginal distributions and posterior distributions, which provide insights for covariate shift scenario. Under the theoretical guarantee and covariate shift assumption, i.e., posterior distributions are unchanged, the covariate shift correction methods aim to mitigate marginal shift via reweighting or alignment. For importance reweighting methods, it directly corrects the shift in original space $\mathcal{X}$ with asymptotic property, where the risk estimation under reweighting distribution is unbiased \wrt the target domain. On the other hand, for transformation-based methods, they aim to learn a transformation $g: X \mapsto Z $ such that the push-forward measures are aligned, i.e., $P_Z = Q_Z$. However, note that $g(\cdot)$ does not necessarily preserve the identical posterior property in representation space, i.e., $P_{Y|Z} = Q_{Y|Z}$ if and only if $g(\cdot)$ is a bijection, which implies the transformation-based methods could not be sufficient to learn unbaised model.

Recently, both empirical study~\cite{long2018conditional,ren2018generalized,gu2021adversarial} and theoretical results~\cite{combes2020domain,federici2021information,stojanov2021domain,kirchmeyer2022mapping} show that learning conditional invariant property via label variable $Y$ is crucial for more general scenario. Therefore, it is necessary to relax the strict deterministic form assumption on $Y$ by considering a `soft' probabilistic form. Specifically, by decomposing the joint distribution as $P_{XY}=P_{X|Y}P_{Y}$, conditional shift correction models aim to learn invariant transformation $g: X \mapsto Z $ such that $P_{Z|Y} = Q_{Z|Y}$. Similarly, label shift correction models~\cite{gu2021adversarial,yan2017mind,zhang2013domain} assume that $P_{Y}\neq Q_{Y}$ and introduce a reweighting strategy to adjust the distribution $P_{Y}$. By integrating the conditional shift and label shift, GLS is developed to consider both shifts on $P_{X|Y}$ and $P_Y$. Recent theoretical results have contributed a lot to the sufficiency of GLS correction by defining conditional discrepancy on predictor~\cite{combes2020domain} or label-wise Wasserstein distance on representations~\cite{kirchmeyer2022mapping}. Different from these efforts, our work focus on both sufficiency and necessity of GLS correction, while deriving more informative generalization bounds with explicit error decomposition.

\textbf{An Information-Theoretic Lower Bound.} 
To provide some insights into the limitation of covariate shift-based, Zhao {\etal}~\cite{zhao2019learning} derive an intuitive lower bound from an information-theoretic perspective, which also implies the impact of label shift is non-negligible.
\begin{theorem}[Lower bound of joint error~\cite{zhao2019learning}]
  \label{thm:zhao_2019_IT-lower-bound}
  Assume that $d_{\JS}(P_Y,Q_Y)\geq d_{\JS}(P_Z,Q_Z)$. Then for any transformation $g: X \mapsto Z$ and hypothesis $h: Z \mapsto Y$, we have
  \begin{equation}
    \label{eq:zhao_2019_IT-lower-bound}
    \varepsilon_P(h\circ g) + \varepsilon_Q(h\circ g) \geq \frac{1}{2} \left[ d_{\JS}(P_Y,Q_Y) - d_{\JS}(P_Z,Q_Z) \right]^2.
  \end{equation}
\end{theorem}
Thm.~\ref{thm:zhao_2019_IT-lower-bound} shows that the marginal adaptation with $P_Z = Q_Z$ is error-prone when label shift exists, i.e., $d_{\JS}(P_Y,Q_Y)>0$. In this case, the R.H.S. of Eq.~\eqref{eq:zhao_2019_IT-lower-bound} is always non-zero, and there will never be a perfect model $h\circ g$ such that the joint error is zero. \textit{However, it is still unclear whether other transformations can alleviate this problem.} For example, matching conditional distributions may also be insufficient, where the lower bound value is still uncertain as both two $d_{\JS}$ terms may be non-zero. In this work, we will give more explicit results which show the \textit{insufficiency of any transformation}.

\subsection{Invariant Representation Learning}
\label{subsec:preliminary_method}
Generally, invariant representation learning focuses on the feature transformation $g$ with different distribution matching properties. Inspired by theoretical results, the invariant representation learning algorithms can be roughly summarized as follows.

Marginal alignment methods usually employ different statistical discrepancies to matching the marginal distributions $P_Z$ and $Q_Z$, where the theoretical guarantee is ensured by covariate shift theory~\cite{ben2010theory,zhang2019bridging}. Typical methods including kernel-based maximum mean discrepancy (MMD)~\cite{gretton2012kernel,long2015learning}, adversarial training with implicit JS divergence~\cite{ganin2016domain,xia2021adaptive} and Wasserstein distance with optimal transport (OT)~\cite{courty2016optimal,li2020enhanced,zhang2019optimal}. As the label distributions are usually imbalanced and shifting in real-world scenarios, theoretical work~\cite{zhao2019learning} shows that the marginal alignment is insufficient and the bias induced by label shift is non-negligible. Since the label shift induces misalignment problem, reweighting alignment improves marginal alignment by reconstructing marginal distribution with weights on labels. Typical methods including reweighting MMD~\cite{yan2017mind}, reweighting adversarial training~\cite{zhang2018importance,stojanov2021domain} and reweighting OT~\cite{gu2021adversarial,redko2019optimal}.

Conditional alignment is proposed to match local structure and mitigate negative transfer. As conditional alignment focuses on distributions $P_{Z|Y=y}$ and $Q_{Z|Y=y}$, it is relatively reliable when label shift exists, and also more accurate than marginal/reweighting methods with global structure matching~\cite{gong2016domain,combes2020domain}. Typical methods including conditional MMD~\cite{ren2016conditional}, adversarial training with conditional information~\cite{long2018conditional} and conditional variants of OT~\cite{luo2021conditional,ren2022buresnet}.

Though invariant representation learning algorithms are sufficient to ensure the transferability in representation space $\mathcal{Z}$, the learned hypothesis may still be biased due to the label shift or suboptimal due to essential trade-off. Specifically, on the one hand, for the limitations on correcting bias/shift, recent works on GLS~\cite{zhang2013domain,combes2020domain,kirchmeyer2022mapping,stojanov2021domain,rakotomamonjy2021optimal} have shown the effectiveness of integrating invariant representation learning with label shift correction. These methods simultaneously align the conditional distributions and apply a reweighting function to the empirical risk optimization of hypothesis $h$, e.g., MMD-based GLS correction~\cite{zhang2013domain,ren2018generalized}, adversarial-based GLS correction~\cite{combes2020domain} and OT-based GLS correction~\cite{kirchmeyer2022mapping,rakotomamonjy2021optimal}. On the other hand, for the limitation in model's optimality, Zhao~\etal~\cite{zhao2022fundamental} analyze the fundamental limits by considering the trade-off between task accuracy and shift invariance, where the suboptimality of existing invariant representation learning methods is theoretically demonstrated. Methodologically, an analytic solution for trade-off is derived under Pareto optimality.

For these advanced invariant represent learning algorithms, the major limitations are the implicit conditional alignment and incomplete theoretical guarantee. In this work, an explicit GLS correction framework is proposed, which also unifies the existing invariant learning algorithms, and provides comprehensive guarantees from both sufficiency and necessity aspects. Besides, compared with the limitation analysis on essential trade-off, our work focuses on the theoretical analysis for limitations on shift correction.

\section{Theoretical Results}
\label{sec:main_results}

\textbf{Motivation.} 
Generally, previous results demonstrate that both conditional shift and label shift are crucial to achieve sufficiently small generalization bound/error, which also implies the essential of dataset shift, i.e., joint distribution matching. In the following, we first show the impossibility of joint distribution matching with invariant transformation $g$ in Sec.~\ref{subsec:theory_limitations_IRL}, which justifies the limitations of invariant representation learning. Further, by introducing the `reweighting source domain', we prove a tighter generalization upper bound and an informative lower bound in Sec.~\ref{subsec:sufficiency_necessity_GLS}, which admit the sufficiency and necessity of GLS correction. Finally, we analyze the dataset shift from the perspectives of Bayes error rate in Sec.~\ref{subsec:hypothesis_bayes_classifier}, which ensures the hypothesis learned with GLS correction is sufficiently close to the optimal classifier. The proofs of theoretical results are provided in \textit{supplementary material}.

The formal definition of GLS is firstly presented by Tachet des Combes~\etal\cite{combes2020domain}, where the main motivation is to relax the strong assumption in label shift, i.e., conditional distributions are identical in input space $\mathcal{X}$. 
\begin{definition}[\cite{combes2020domain}]
  \label{def:GLS_combes}
  The representation $Z=g(X)$ satisfies GLS if $P_{Z|Y}=Q_{Z|Y}$.
\end{definition}
Intuitively, the GLS considers the identical assumption on intermediate space $\mathcal{Z}$, which implies the invariant representation learning on conditional distribution. 

Motivated by this, we further consider a general definition that can mathematically unify the existing invariant representation learning methods. Focusing on the properties of learner $g$, the rigorous definition of representation transformation with probability property (i.e., pushforward distribution) can be presented as follows.

\begin{definition}[Invariant transformations]
  \label{def:invariant-transformation}
  Let $g: X \mapsto Z $ be a measurable mapping (transformation) and $P_{Z}$ (resp. $Q_{Z}$) be the pushforward of $P_{X}$ (resp. $Q_{X}$) via $g(\cdot)$.
  \begin{enumerate}[label=(\alph*), ref=\ref*{def:invariant-transformation}\alph*]
    \item A transformation $g$ is called marginal invariant (denoted as $g_{\mar}$) if it satisfies that $P_{Z}=Q_{Z}$; \label{def:margin-invariant-transformation}
    \item A transformation $g$ is called conditional invariant (denoted as $g_{\con}$) if it satisfies that $P_{Z|Y}=Q_{Z|Y}$ \label{def:condition-invariant-transformation}
  \end{enumerate}
\end{definition}
Since the down-stream tasks (e.g., classification) are usually considered in latent space $\mathcal{Z}$, we define the hypothesis space as $\mathcal{H}=\{h| h: Z\mapsto Y\}$ hereinafter. By substituting $g_{\mar}$ into Thm.~\ref{thm:zhao_2019_IT-lower-bound} by Zhao~\etal~\cite{zhao2019learning}, the impossibility result for marginal invariant representation learning can be directly deduced.
\begin{remark}
  \label{rem:imposs_marginal}
  Assume that label shift exists. For any hypothesis $h: Z \mapsto Y$, we have $\varepsilon_P(h\circ g_{\mar}) + \varepsilon_Q(h\circ g_{\mar}) ~>0~$.
\end{remark}

\begin{figure*}[t]
  \centering
  \includegraphics[width=0.99\linewidth,trim=30 15 20 30,clip]{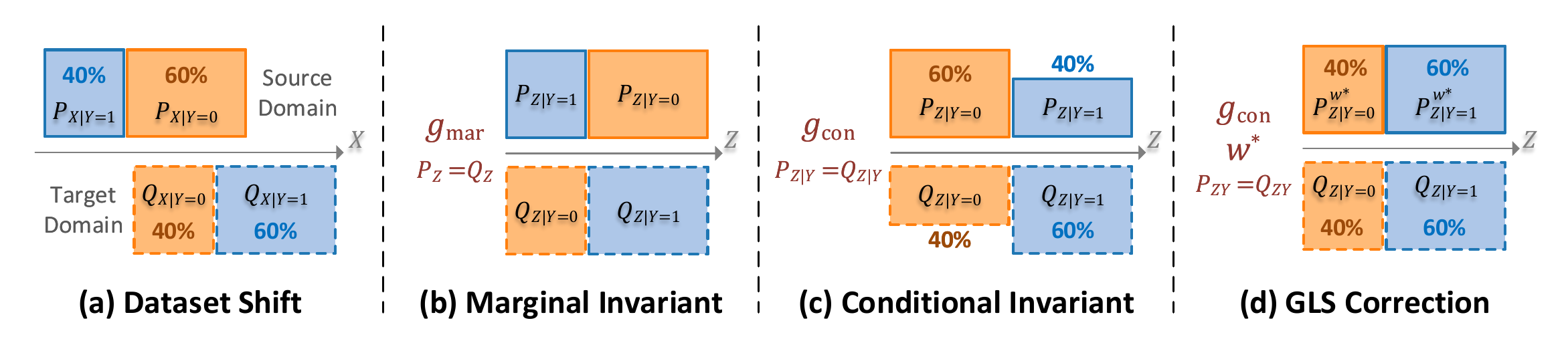}
  \caption{Illustration of the invariant representation learning and dataset shift correction in $\MBB{R}^1$. (a) The dataset shift exists as $P_{X|Y}\neq Q_{X|Y}$ and $P_Y\neq Q_Y$, where $p_Y=[\textcolor{fig_orange}{0.6};\textcolor{fig_blue}{0.4}]$ and $q_Y=[\textcolor{fig_orange}{0.4};\textcolor{fig_blue}{0.6}]$. (b) Marginal invariant transformation $g_{\mar}$ may misalign the conditional distributions. (c) Conditional invariant transformation $g_{\con}$ is still insufficient to align the joint distributions since label shift leads to different proportions of aligned conditional distributions. (d) GLS correction is sufficient to address dataset shift with the $w^*$ reweighting proportions of conditional distributions.}
  \label{fig:dataset_shift_correction}
\end{figure*}

From the perspective of joint error, this remark explicitly shows the limitations of marginal invariant transformation $g_{\mar}$, where no hypothesis can be optimal across domains \textit{even if the datasets are separable}.

\subsection{Insufficiency of Invariant Representation Learning}
\label{subsec:theory_limitations_IRL}

In this section, we will show the insufficiency of invariant representation learning under \textit{label shift} scenario. Recently, some methods try to enhance the transferability by learning the marginal invariant and conditional invariant properties simultaneously, i.e., incorporating covariate alignment and conditional alignment into one model. However, the lack of theoretical understanding and interpretation lead to uncertainty in learning those properties. Here we first introduce an important concept for theoretical analysis, and then provide an impossibility result for achieving $g_{\mar}$ and $g_{\con}$ simultaneously.
\begin{definition}[Linear independence of functions]
  \label{def:function_linearly_independent}
  Assume that $\MC{F}=\{f_y:\MC{X}\rightarrow \MBB{R}~|~ y\in\MC{Y}\}$ be a set of real-valued functions. Then the functions in $\MC{F}$ are linearly independent if there does not exist non-zero function $t(\cdot):\MC{Y}\rightarrow \MBB{R}$ for linear combination $f(X) \triangleq \int_\MC{Y} f_y(X) t(y) \diff y$ such that
  \begin{equation*}
    f(X) ~ \equiv ~  0.
  \end{equation*}
\end{definition}

Intuitively, the linear independence implies that each function in $\mathcal{F}$ is ``unique'', i.e., any function $f\in\mathcal{F}$ cannot be linearly expressed by other rest functions in $\mathcal{F}$, which is analogous to the independence in vector space.

\begin{restatable}{proposition}{ProImpossMarConSimul}
  \label{pro:imposs-margin-condition-simultaneously}
  Assume that label shift exists and the conditional distributions in set $\mathcal{P}_{Z|Y} \triangleq \{P_{Z|Y=y}~|~ y\in\MC{Y}\}$ are linearly independent of each other. There is no transformation $g$ such that $P_{Z}=Q_{Z}$ and $P_{Z|Y}=Q_{Z|Y}$, which implies that $g_{\mar}\neq g_{\con}$ always holds.
\end{restatable}
Prop.~\ref{pro:imposs-margin-condition-simultaneously} shows that the marginal invariant transformation $g_{\mar}$ and conditional invariant transformation $g_{\con}$ are mutually exclusive when label shift exists. For example, by considering the classification scenario, an intuitive justification for this impossibility is that the different label proportions across domains lead to different importance (i.e., $P_Y\neq Q_Y$) of local structures (i.e., conditional distributions). Then, if the clusters are matched, the global structures (i.e., marginal distributions) reconstructed by local structures will be different across domains, i.e., $P_Z= \MBB{E}_{P_Y} [P_{Z|Y}] = \MBB{E}_{P_Y} [Q_{Z|Y}] \neq \MBB{E}_{Q_Y} [Q_{Z|Y}]= Q_Z$, and vice versa. Note that the independence assumption on $\mathcal{P}_{Z|Y}$ is reasonable and generally holds~\cite{kirchmeyer2022mapping,gong2016domain,combes2020domain}, which is necessary to ensure the prediction information of $Y$ is preserved by transformation $g$. For example, for any $y_0\neq y_1 \in \mathcal{Y}$, this independence assumption implies that the conditional distributions (e.g., $P_{Z|Y=y_0}$ and $P_{Z|Y=y_1}$) are not totally overlapped. This is reasonable since if they are totally overlapped, it will be impossible to identify them and the conditional distributions no longer depend on $Y$, which implies the discriminant information of $Y$ is lost. Therefore, it can be concluded from Prop.~\ref{pro:imposs-margin-condition-simultaneously} that a model is impossible to simultaneously achieve marginal invariant and conditional invariant properties, unless a basic discriminability for task (i.e., independence in $\mathcal{P}_{Z|Y}$) is lost.

As discussed in Thm.~\ref{thm:zhao_2019_IT-lower-bound}, marginal alignment cannot ensure an optimal hypothesis across domains. This result reveals the limitation of the marginal invariant transformation $g_{\mar}$. In the next, we provide a stronger result for the insufficiency of invariant representation learning, which implies that \textit{any transformation} $g$, including $g_{\mar}$ and $g_{\con}$, are error-prone when label shift exists.
\begin{restatable}[Impossibility of dataset shift correction]{proposition}{ThmImpossJointMatch}
  \label{prop:imposs-joint-matching}
  If label shift exists, there does not exist a transformation $g: X \mapsto Z $ that corrects the dataset shift, i.e., $P_{ZY}\neq Q_{ZY}$ for any $g$.
\end{restatable}
Prop.~\ref{prop:imposs-joint-matching} suggests that any transformation $g$ is insufficient to tackle the dataset shift problem, i.e., the shift of joint distributions $P_{XY}\neq Q_{XY}$. This impossibility result is also intuitively shown as Fig.~\ref{fig:dataset_shift_correction}. It implies that invariant representation learning is not sufficient for the invariant risk estimation across domains, i.e., $\varepsilon_P (h)=\varepsilon_Q (h)$ for any $h\in\mathcal{H}$. \textit{Moreover, it proves that the label shift cannot be implicitly addressed by transformation on $Z$ while an explicit correction on $Y$ is always necessary.} 

On the other hand, from the perspectives of risk values of models, we can conclude more intuitive results as follows. For $g_{\mar}$, it is clear that it still cannot address the bias in risk estimation and always induces non-zero joint error as discussed in Rem.~\ref{rem:imposs_marginal}. Thus, it is natural to consider the property of $g_{\con}$ {\wrt} risk estimation.

\begin{proposition}
  \label{prop:conditional_risk_property}
  Given transformation $g$, denote the set of overlapped clusters $\Omega_1=\{y~|~ \supp(p_{Z|Y=y}) \cap \supp(p_{Z|Y=y'}) \neq \varnothing, y'\in \mathcal{Y}\}$.
  \begin{enumerate}[label=(\alph*)]
    \item If $\Omega_1 \neq \varnothing$ holds for $g_{\con}$, then $\varepsilon_P(h\circ g_{\con})>0$ and $\varepsilon_Q(h\circ g_{\con})>0$ also hold for any hypothesis $h$.
    \item If $\Omega_1 = \varnothing$ holds for $g_{\con}$, then there always exists hypothesis $h$ such that $\varepsilon_P(h\circ g_{\con})=\varepsilon_Q(h\circ g_{\con})=0$.
  \end{enumerate}
\end{proposition}

Generally, Prop.~\ref{prop:conditional_risk_property} shows the possible failure and successful scenarios of $g_{\con}$. \textit{(a)} if the dataset is not separable in representation space, i.e., $\Omega_1\neq \varnothing$ holds for $g_{\con}$, the conditional invariant transformation $g_{\con}$ is still insufficient to ensure the existence of perfect learner $h$ on target $Q$; besides, since $g_{\con}$ is insufficient to address dataset shift as shown in Prop.~\ref{prop:imposs-joint-matching}, the risk estimation could still be biased, i.e., $|\varepsilon_P(h\circ g_{\con})-\varepsilon_Q(h\circ g_{\con})|\geq 0$. \textit{(b)} $g_{\con}$ could be able to achieve consistent risk estimation if the dataset is separable, i.e., there exists $g$ such that $\Omega_1= \varnothing$. Combining these two scenarios, it is clear that $\varepsilon_P(h\circ g_{\con})+\varepsilon_Q(h\circ g_{\con})\geq 0$ always holds, where the equality holds if and only if the dataset is separable. Since it is usually hard to guarantee the existence of perfect learner in real-world application, $g_{\con}$ will always induce insufficiency and (even) bias in risk estimation. Intuitively, since the risk estimation is mathematically formulated as the expectation over joint distribution, the bias could be induced by the disagreement on the importance of aligned clusters, i.e., $P_Y\neq Q_Y$.

Overall, the results imply that distribution matching via transformation, e.g., $g_{\mar}$ and $g_{\con}$, is impossible to fully resolve the dataset shift problem $P_{ZY}\neq Q_{ZY}$. Moreover, Prop.~\ref{pro:imposs-margin-condition-simultaneously} shows that $g_{\mar}$ usually leads to the misaligned cluster structures (i.e., conditional distributions); Prop.~\ref{prop:conditional_risk_property} further show that $g_{\con}$ also cannot ensure an ideal representation space for the hypothesis learning on target domain. Thus, the results suggest that both $g_{\mar}$ and $g_{\con}$ are also insufficient to eliminate the bias in risk estimation, and a correction on label shift $P_{Y}\neq Q_{Y}$ is necessary and potentially effective.

\subsection{Sufficiency of GLS Correction}
\label{subsec:sufficiency_necessity_GLS}

Hopefully, though invariant representation learning is not sufficient for dataset shift correction, the conditional distribution matching at least preserves the intrinsic local structure for downstream task. Compared with previous model with pair $(g,h)$, we extend the model under dataset shift as a triplet $(g,h,w)$ by introducing the label weight $w$, and then define the ideal model for GLS.

\begin{definition}
  \label{def:reweighting_and_GLS-correction}
  Let $w: \MC{Y}\rightarrow \MBB{R}_{+}$ be a function such that $wp_Y$ is a PDF on $\MC{Y}$.
  \begin{enumerate}[label=(\alph*), ref=\ref*{def:invariant-transformation}\alph*]
    \item Given a source distribution $P$. A $w$-weighting source distribution, which is denoted by $P^w_{ZY}$, is defined as $p^w_Y = wp_Y$ and $p^w_{Z|Y} = p_{Z|Y}$; \label{def:w-reweighting-distribution}
    \item Denote the learning model under dataset shift as ($g,h,w$) and optimal weight as $w^*\triangleq  \frac{q_Y}{p_Y}$. A model is called GLS correction model if it satisfies that $g=g_{\con}$ and $w=w^*$.  \label{def:GLS-correction}
  \end{enumerate}
\end{definition}

\textbf{Sufficiency of GLS Correction.}
The importance weighting on source domain has been empirically proved to be effective in label shift and GLS scenarios~\cite{combes2020domain,gong2016domain,lipton2018detecting}. Based on the definition of $w$-reweighting distribution, we next present an informative generalization upper bound which ensures the sufficiency of GLS correction model ($g_{\con},h,w^*$) for successful knowledge transfer. 
\begin{restatable}[Sufficiency of GLS correction]{theorem}{ThmSufficiencyGLS}
  \label{thm:GLS-upper-bound}
  Assume that the loss function $\ell$ is bounded (with constant $M$). Then for any learning model ($g,h,w$),
  \begin{equation}
    \label{eq:GLS-upper-bound}
    \begin{array}{l}
    \underbrace{| \varepsilon_{P^w} - \varepsilon_Q |}_{\textbf{dataset shift}}
    \leq 2M \big[ 
    \underbrace{d_{\TV}(P^w_{Y},Q_{Y})}_{\textbf{label shift}} \\
    + 
    \underbrace{ \min \{ \MBB{E}_{P^w_Y} [d_{\TV}(P^w_{Z|Y},Q_{Z|Y})], \MBB{E}_{Q_Y} [d_{\TV}(P^w_{Z|Y},Q_{Z|Y})] \}}_{\textbf{conditional shift}}
     \big].
    \end{array}
  \end{equation}
  Especially, for GLS correction model ($g_{\con},h,w^*$), we have
  \begin{equation*}
    \label{eq:GLS-correction_optimal_identity}
    P^{w^*}_{ZY} = Q_{ZY} 
    \quad ~~ \mathrm{and} ~~ \quad
    \varepsilon_{P^{w^*}}(h\circ g_{\con}) = \varepsilon_Q(h\circ g_{\con}).
  \end{equation*}
\end{restatable}

Thm.~\ref{thm:GLS-upper-bound} is consistent with recent works on GLS correction models~\cite{zhang2013domain,gong2016domain,combes2020domain,kirchmeyer2022mapping,rakotomamonjy2021optimal}. Specifically, the conditional shift term will be sufficiently small by learning conditional invariant transformation $g_{\con}$, while the label shift term induced by the intrinsic difference cannot be directly mitigated by transformation $g$. Hopefully, the label shift term can be dominated by the weight function $w$ if the model is trained on the reweighting source domain $P^{w}$. Therefore, a proper weight $w$ can adjust the label proportion in risk estimation and reduce label shift effectively. Finally, an ideal GLS correction model ($g_{\con},h,w^*$) is sufficient to address the dataset shift problem as shown in Fig.~\ref{fig:dataset_shift_correction}.

\textbf{Implications Compared with Existing Bounds}. 
Moreover, Thm.~\ref{thm:GLS-upper-bound} can be compared to recent advances~\cite{zhao2019learning,combes2020domain,li2021learning,kirchmeyer2022mapping} on generalization error bounds. Compared with the recent GLS-based generalization upper bounds for 0-1 loss function~\cite{combes2020domain,kirchmeyer2022mapping}, Thm.~\ref{thm:GLS-upper-bound} contributes to a general application scenario and ensures informative conclusions: 1) it holds for more general cases with any bounded loss functions $\ell$ and output space, e.g., continuous output for regression and discrete output for classification; 2) it is more reliable when conditional shift is severe. Specifically, based on predictor $\hat{Y}=h\circ g(X)$, Tachet des Combes \etal~\cite[Thm. 3.1]{combes2020domain} also provides an upper bound that shows the same decomposition form in Thm.~\ref{thm:GLS-upper-bound}:
\begin{equation*}
  \label{eq:combes_upper_bound}
  |\varepsilon_P-\varepsilon_Q|\leq 2\cdot \mathrm{BER}_s(\hat{Y}||Y) d_{\TV}(P_Y,Q_Y) + 2(K-1)\Delta_{\mathrm{CE}}(\hat{Y})
\end{equation*}
where $K=|\mathcal{Y}|$, $\Delta_{\mathrm{CE}}(\hat{Y}) = \underset{y\neq y'}{\max}~|p_{\hat{Y}=y'|Y=y}-q_{\hat{Y}=y'|Y=y}|$ and $\mathrm{BER}_s(\hat{Y}||Y)=\underset{y\in\mathcal{Y}}{\max}~p_{\hat{Y}\neq y | Y=y}$. By considering representation $Z$ as prediction $\hat{Y}$ (this is reasonable since it is for theoretical comparison) and loss function as 0-1 loss, i.e., $M=1$, we can compare the bound above with Eq.~\eqref{eq:GLS-upper-bound}. Since $\mathrm{BER}_s(\hat{Y}||Y)\leq 1$, the label discrepancy term above is generally smaller than that in Eq.~\eqref{eq:GLS-upper-bound}. In contrast, if $K> 2$, which generally holds for real-world application, it is clear that 
\begin{align*}
  &~2(K-1)\Delta_{\mathrm{CE}}(\hat{Y}) \\
  >& ~K \max_{y\neq y'}~|p_{\hat{Y}=y'|Y=y}-q_{\hat{Y}=y'|Y=y}|  \\
  \geq &  ~2 \max_{y\in \mathcal{Y}}~ d_{\TV}(P_{\hat{Y}|Y=y},Q_{\hat{Y}|Y=y}) \\
  \geq & ~2 \max\{ \mathbb{E}_{P_Y}[d_{\TV}(P_{\hat{Y}|Y},Q_{\hat{Y}|Y})], \mathbb{E}_{Q_Y}[d_{\TV}(P_{\hat{Y}|Y},Q_{\hat{Y}|Y})] \}.
\end{align*}
Thus, the conditional discrepancy term above is strictly larger than that in Eq.~\eqref{eq:GLS-upper-bound}. Further, $2(K-1)\Delta_{\mathrm{CE}}(\hat{Y})$ tends to be loose as $K$ increase, since the gap between $2(K-1)$ and $K$ will be enlarged and the scale of omitted $(K-2) \Delta_{\mathrm{CE}}(\hat{Y}) $ will be increased. In conclusion, compared with the bound in literature \cite{combes2020domain}, our bound is more compact when conditional discrepancy term dominates the error, and vice versa.

Besides, compared with the error decomposition by Zhao~\etal~\cite{zhao2019learning} and Li~\etal~\cite{li2021learning}, Thm.~\ref{thm:GLS-upper-bound} can be similarly extended to the covariate shift and concept shift scenario by decomposing the joint distribution as $P_{ZY}=P_{Y|Z}P_{Z}$, then the upper bound consists of covariate shift term $d_{\TV}(P^w_{Z},Q_{Z})$ and concept shift term $d_{\TV}(P^w_{Y|Z},Q_{Y|Z})$. However, from the perspective of representation learning, the decomposition form in GLS ensures that the error factor that can be controlled by transformation $g$ is sufficiently expressed in single term, i.e., the conditional discrepancy. Differently, the decomposition form based on $P_{ZY}=P_{Y|Z}P_{Z}$ induces two discrepancy terms that are both related with $g$ and the invariant properties on these two terms can also be contradicted (if label shift exists) from the law of total probability $P_Y = \int_{\mathcal{Z}} P_{Y|Z=z} \cdot p_{Z=z} \diff z$. Besides, the concept shift term intuitively implies the labeling rule, which is generally intractable in practical learning process. Therefore, the form of GLS is more preferable in the view of representation learning, since it characterize the shift correction via two independent learners $g$ and $w$, which provide explicit connection between the learners and generalization upper bound terms.

\subsection{Necessity of GLS Correction}
\label{subsec:necessity_GLS}

\textbf{Necessity of GLS Correction.}
Recent works~\cite{federici2021information,zhao2019learning} also analyze the lower bound of the generalization error from the perspectives of information theory. These works reveal the limitations of the marginal invariant transformation (i.e., Thm.~\ref{thm:zhao_2019_IT-lower-bound}) and provide insights into understanding dataset shift with the discrepancy between concepts on representation space, i.e., posterior distributions $P_{Y|Z}$. Inspired by the idea of information theory, we next show the discrepancy between concepts across domains is lower bounded by the amount of domain-specific information preserved in representations, which implies the transferability of concepts.

Following the same notations, to characterize the domain information, we first define the domain as a random variable $D$ which takes its values from $\MC{D}= \{s,t\}$. Then the notation of distribution is extended as $R^w_{XYD}$ which satisfies that $R^w_{XY|D=s} = P^w_{XY}$ and $R^w_{XY|D=t} = Q_{XY}$. The notations {\wrt} $R^w_{ZYD}$ are similar. Denote $a=r^w_{D=t}$ and $1-a =r^w_{D=s}$, where $r^w_{D=s}$ and $r^w_{D=t}$ represent the masses (i.e., probabilities or proportions) of the source and target domains, respectively. Let $I(\cdot~;\cdot|\cdot)$ be the conditional mutual information~\cite{cover1999elements}.

\begin{restatable}[Necessity of GLS correction]{theorem}{ThmNecessityGLS}
  \label{thm:GLS-lower-bound}
  For any learning model ($g,h,w$), let $a=r^w_{D=t}$, $b = \min\{r^w_{D=s},r^w_{D=t}\}$ and $\gamma(z)=\max \{p^w_{Z},q_{Z}\}$. Assume that $d_{\JS,a}(P^w_Y,Q_Y) \geq d_{\JS,a}(P^w_Z,Q_Z)$, then
  \begin{equation}
    \label{eq:GLS-lower-bound}
    \underbrace{\int_\MC{Z} \gamma(z) d_{\JS}(P^w_{Y|Z=z},Q_{Y|Z=z}) \diff z}_{\textbf{disagreement on posterior distributions}} 
     \geq 
    \underbrace{\frac{1}{2(1-b)} I^w(Z;D|Y)}_{\textbf{GLS correction}} 
    ,
  \end{equation}
  where
  \begin{align}
      I^w(Z;D|Y)&=\MBB{E}_{R^w_Y} \big[ \MBB{E}_{R^w_D} [d_{\KL}(R^w_{Z|YD} \| R^w_{Z|Y})] \big] \nonumber \\
      &= \MBB{E}_{R^w_Y} \big[ d_{\JS,a}(P^w_{Z|Y},Q_{Z|Y}) \big] \label{eq:GJS-divergence}.
  \end{align}
\end{restatable}

Thm.~\ref{thm:GLS-lower-bound} shows that conditional invariant transformation $g_{\con}$ is necessary to reduce the disagreement on the labeling rules across domains, which directly characterizes the precision when transferring a model on source domain to target domain. From the information-theoretic view, the amount of information between covariate $X$ and domain variable $D$ should be minimized by encoding $X$ to the representation $Z$. Moreover, the information minimization procedure is necessarily conditioned on the label variable $Y$, i.e., $I^w(Z;D|Y)$, which prevents the loss of discriminant information during encoding invariant representations. From the statistical view, the mutual information $I^w(Z;D|Y)$ represents the conditional dependence between representation $Z$ and domain $D$ given label $Y$. Then the zero mutual information will admit conditional independence between $Z$ and $D$, which is equivalent to $R^w_{X|YD=s} = R^w_{X|YD=t}$. From Eq.~\eqref{eq:GJS-divergence}, it is clear that $I^w(Z;D|Y)=0$ if and only if $P^w_{Z|Y}=Q_{Z|Y}$, which implies the conditional invariant model ($g_{\con},h,w$) is necessary for adaptation.

Besides, Tachet des Combes~\etal~\cite[Thm. 3.4]{combes2020domain} also provide a necessity result from the perspective of joint error, i.e., the conditional discrepancy serves as the lower bound of cross-domain joint error. Differently, Thm.~\ref{thm:GLS-lower-bound}, focuses on the necessity for shift correction, i.e., the minimized disagreement across environments. Consequently, Tachet des Combes~\etal show that $g_{\con}$ is necessary to ensure that $\varepsilon_P+\varepsilon_Q$ can be arbitrary small and Thm.~\ref{thm:GLS-lower-bound} is necessary to ensure that the cross-domain labeling rules $P_{Y|Z}$ and $Q_{Y|Z}$ can be arbitrary close.

Recall that a mild assumption on label distribution discrepancy and representation distribution discrepancy is made in Thm.~\ref{thm:GLS-lower-bound}, i.e., $d_{\JS,a}(P^w_Y,Q_Y) \geq d_{\JS,a}(P^w_Z,Q_Z)$. Fortunately, it is generally valid, where the reasons can be concluded from the following two apsects. Intuitively, this assumption is reasonable since the discrepancy $d_{\JS,a}(P^w_Z,Q_Z)$ between representations is explicitly optimized by the transformation $g$. Theoretically, we present the next lemma to provide a sufficient condition for this assumption. 
\begin{restatable}{lemma}{LemAssumpJSD}
  \label{lem:assumption_on_JSD}
  For any weight function $w$, if $g$ is conditional invariant, i.e., $g=g_{\con}$, then
  \begin{equation*}
    I^w(Y;D) \geq I^w(Z;D),
  \end{equation*}
  which is equivalent to $d_{\JS,a}(P^w_Y,Q_Y) \geq d_{\JS,a}(P^w_Z,Q_Z)$.
\end{restatable}
Lem.~\ref{lem:assumption_on_JSD} proves that the assumption $d_{\JS,a}(P^w_Y,Q_Y) \geq d_{\JS,a}(P^w_Z,Q_Z)$ holds when $g=g_{\con}$. Moreover, it provides a new insight into understanding the dataset shift when $g$ is conditional invariant. Specifically, Lem.~\ref{lem:assumption_on_JSD} shows that the dependence between $Z$ and $D$ is bounded by the dependence between $Y$ and $D$ from the view of mutual information. Intuitively, when $g=g_{\con}$, the aligned label distribution is sufficient to ensure marginal invariant property, i.e., $P^w_{Y}=Q_{Y}\Longrightarrow P^{w}_Z = Q_Z$. Therefore, it implies that the information between representations and domains is only dominated by the ``amount of information'' between $Y$ and $D$ (i.e., label shift) when conditional invariant property is ensured.

A similar setting is also studied by Zhao~\etal~\cite[Prop. 3.2]{zhao2019conditional}, while they focuses on the fairness of model with binary output. Specifically, they consider the marginal discrepancies on label $Y$ and prediction $\hat{Y}$, and show that $d_{\TV}(P_{\hat{Y}},Q_{\hat{Y}})\leq d_{\TV}(P_Y,Q_Y)$. Compared with this result, the major differences are that Lem.~\ref{lem:assumption_on_JSD} focuses on the label $Y$ and representation $Z$, and is valid for any output space.

\textbf{Implications for Invariant Risk Minimization}. 
As a closely related area that also focuses on the generalization on different environments, invariant risk minimization (IRM) framework~\cite{arjovsky2019invariant,li2022invariant} has received extensively attention. Mathematically, IRM aims to learn transformation $g$ and predictor $h$ such that $h\circ g$ consistently minimizes the error on multiple changing environments. For example, in distribution shift scenario, IRM can be formulated as~\cite{arjovsky2019invariant}
\begin{equation*}
  \label{eq:IRM_objective}
  \begin{array}{l}
  \underset{h, g}{\min}~ \varepsilon_P(h\circ g) +  \varepsilon_Q(h\circ g) \\
  \textrm{s.t.} ~~ h \in \underset{h'}{\arg \min} ~\varepsilon_P(h'\circ g) ~~\text{and}~~
  h \in \underset{h'}{\arg \min} ~\varepsilon_Q(h'\circ g).
  \end{array}
\end{equation*}

Fortunately, the necessity results mentioned before can also provide insights for IRM. It is clear that the results of small joint error (i.e., \cite[Thm. 3.4]{combes2020domain}) and invariant labeling rule (i.e., Thm.~\ref{thm:GLS-lower-bound}) can be directly connected to the objective and constraints in IRM, which shows $g_{\con}$ is also necessary for IRM. Beyond the necessity of IRM, Li~\etal~\cite{li2021learning} provide insights for the sufficiency by showing the original IRM framework may induce pseudo-invariant representations, i.e., the components that are independent not only of environment but also of labels, and regularization for filtering label-independent components (e.g., noise and color) are the key for sufficiency.

In conclusion, we have proved that GLS correction is indeed sufficient (i.e., Thm.~\ref{thm:GLS-upper-bound}) and necessary (i.e., Thm.~\ref{thm:GLS-lower-bound}) for knowledge transfer. The upper-bound Eq.~\eqref{eq:GLS-upper-bound} in sufficient theorem provides an explicit decomposition of risk estimation error, which directly guarantees the effectiveness of GLS correction algorithms. The lower-bound Eq.~\eqref{eq:GLS-lower-bound} in necessity theorem implies that the minimum generalization ability of model is dominated by label-conditioned information between representations and domains. 

\subsection{Optimal Hypothesis: An Agnostic Setting}
\label{subsec:hypothesis_bayes_classifier}

As discussed in Sec.~\ref{sec:preliminary}, previous works~\cite{ben2010theory,cortes2019adaptation,zhao2019learning} usually consider a deterministic form of $Y$, i.e., $Y$ is decided by a labeling rule $f(\cdot)$. Then these works reach a consistent conclusion that the existence of a cross-domain optimal hypothesis is crucial for the successful adaptation, i.e., a joint optimal hypothesis $h^*$ with the lowest error in both domains. However, by considering $Y$ as deterministic form, this term usually serves an intractable constant in the generalization upper bound since optimal joint error of $h^*$ is hard to evaluate. To provide some insights into understanding domain-specific labeling rules and optimal hypotheses, we further analyze GLS correction from the perspectives of Bayes classifier and Bayes error rate. 

\begin{definition}[Bayes Classifier and Bayes Error Rate~\cite{hastie2009elements}]
  \label{def:bayes_classifier_bayes_rate}
  Let $Z$ be $d$-dimensional random variable with mutually independent entries $Z_1,Z_2,\ldots,Z_d$ given $Y$, and $K=|\MC{Y}|$ be the cardinality of $\MC{Y}$. The Bayes classifier over a given distribution $P_{ZY}$ is defined as
  \begin{equation}
    \label{eq:def_bayes_classifier}
    f_P(z) \triangleq  \mathop{\arg \max}_{k\in [K]} ~ p_{Y|Z}(Y=k|Z=z).
  \end{equation}
  Then the Bayes error rate with 0-1 loss function $\ell_{01}(\cdot,\cdot)$ is a constant defined by
  \begin{align}
      \varepsilon^{\mathrm{Bayes}}_P &\triangleq \varepsilon_P(f_P) \nonumber \\
      &= \MBB{E}_{P_Z} \Big[ \sum_{k\in [K]} \ell_{01}(f_P(Z),k) \cdot p_{Y|Z}(Y=k|Z) \Big] . \label{eq:def_bayes_rate}
  \end{align}
\end{definition}

\begin{restatable}[\cite{hastie2009elements}]{lemma}{LemOptimalityBayes}
  \label{lem:optimality_of_Bayes_classifier}
  Denote the function class of classifiers as $\MC{F} = \{ f~|~ f: \MC{Z} \rightarrow \MC{Y} \}$. Given a distribution $P_{ZY}$, Bayes classifier $f_P$ is the optimal classifier which has the minimum error rate, i.e., Bayes error rate $\varepsilon^{\mathrm{Bayes}}_P=\mathop{\min}_{f\in \MC{F}} ~\varepsilon_P(f)$.
\end{restatable}
Lem.~\ref{lem:optimality_of_Bayes_classifier} shows that the Bayes classifier is the optimal classifier over the function class $\MC{F}$ and the Bayes error rate is the lowest achievable error rate over $\MC{F}$. Note that $Y$ is not necessarily deterministic in our analysis, so the Bayes classifier can be taken as the optimal hypothesis if the hypothesis class $\MC{H}$ is sufficiently rich, i.e., they are equivalent if $\MC{H} = \MC{F}$. Formally, the gap between error and Bayes error is called \textit{excess risk}. Based on excess risk, Zhang~\etal~\cite{zhang2021quantifying} has provided a comprehensive study on the connection between cross-domain excess risks and model's transferability, where the main conclusions show that 1) estimating transferability is no harder than estimating errors on both domains; 2) the generalization error can be characterized via excess risk instead of previous discrepancy measure. Thus, these results provide the understanding of transferability via excess risk. Inspired by the idea of excess risk, we focus on a different aspect, i.e., the connection between invariant representation learning and excess risk. Specifically, we focus on the problem that can the source trained hypothesis (which is learnable in $\mathcal{H}$) be sufficiently close to the target optimal predictor (which is implicit and intractable in $\mathcal{F}$).

In the next, we first show that the risk of joint optimal hypothesis $h^*$ can be sufficiently small with a proper learning model $(g,h,w)$.
\begin{restatable}{proposition}{ProDiscrepancyBayesRate}
  \label{pro:discrepancy_between_bayes_rate}
  Assume that loss function $\ell$ is bounded (with constant $M$). For any learning model ($g,h,w$), let $f_{P^w}$ and $f_Q$ be the Bayes classifiers on $P^w$ and $Q$, respectively. Then
  \begin{align}
    |\varepsilon^{\mathrm{Bayes}}_{P^w} - \varepsilon^{\mathrm{Bayes}}_Q |
    & =
    |\varepsilon_{P^w}(f_{P^w}\circ g) - \varepsilon_Q(f_Q\circ g) | \nonumber \\ 
    & \leq 2M d_{\TV}(P^w_{ZY},Q_{ZY}). \label{eq:discrepancy_between_bayes_rate}
  \end{align}
  Especially, for GLS correction model ($g_{\con},h,w^*$), we have $f_{P^{w^*}}= f_Q$ and $\varepsilon^{\mathrm{Bayes}}_{P^{w^*}} = \varepsilon^{\mathrm{Bayes}}_Q$.
\end{restatable}

\begin{remark}
  \label{rem:GLS_from_Bayes}
  On the one hand, Prop.~\ref{pro:discrepancy_between_bayes_rate} shows that a GLS correction model ($g_{\con},h,w^*$) is sufficient to ensure the equivalent optimal classifiers for both domains, i.e., $f_{P^{w^*}}= f_Q$. On the other hand, as the Bayes is decided by the posterior distribution as Eq.~\eqref{eq:def_bayes_classifier}, Thm.~\ref{thm:GLS-lower-bound} admits that ($g_{\con},h,w^*$) is necessary for the existence of a joint optimal classifier, i.e., $f^*=f_{P^{w^*}}= f_Q$.
\end{remark}

\begin{remark}
  \label{rem:GLS_joint_optimnal_hypothesis}
  Note that the optimal classifier is not necessarily a hypothesis since $\MC{H}\subseteq \MC{F}$. Generally, we learn an optimal hypothesis in $\MC{H}$ as $h^*_{P^w} = \min_{h\in \MC{H}} ~\varepsilon_{P^w}(h)$. Then the model $(g,h^*_{P^w},w)$ serves as a good approximation for the optimal one $(g,f_{P^{w}},w)$. By combining the GLS correction and Prop.~\ref{pro:discrepancy_between_bayes_rate}, it is straightforward to conclude that for optimal source hypothesis $h^*_{P^{w}}$ with any pair $(g,w)$,
  \begin{align}
    &~|\varepsilon_{P^{w}}(h^*_{P^{w}}\circ g) -  \varepsilon^{\mathrm{Bayes}}_Q| \nonumber \\
    \leq&~ | \varepsilon_{P^{w}}(h^*_{P^{w}}\circ g) - \varepsilon^{\mathrm{Bayes}}_{P^w} | 
    +
    2M d_{\TV}(P^{w}_{ZY},Q_{ZY}). \label{eq:hypothesis_error_Bayes}
  \end{align}
  Especially, optimal hypothesis on source $P^{w^*}$ with GLS correction $(g_{\con},w^*)$ can be sufficiently close to the target Bayes classifier $f_Q$, i.e., $|\varepsilon_{P^{w^*}}(h^*_{P^{w^*}}\circ g_{\con}) -  \varepsilon^{\mathrm{Bayes}}_Q| \leq |\varepsilon_{P^{w^*}}(h^*_{P^{w^*}}\circ g_{\con}) -  \varepsilon^{\mathrm{Bayes}}_{P^{w^*}}|$, which ensures the existence of joint optimal classifier/hypothesis across domains.
\end{remark}

In conclusion, the results from the perspective of Bayes error show that 1) though the cross-domain labeling rules may be different in covariate space, they can be learned to be equivalent via GLS correction; 2) for any algorithm with specific hypothesis space $\mathcal{H}$, its optimal hypothesis on reweighting source domain, i.e., $h^*_{P^{w^*}}$, is sufficient to be close to the target optimal classifier and effectiveness on the target domain, where the error is dominated by source risk minimization and GLS correction, i.e., Eq.~\eqref{eq:hypothesis_error_Bayes}.

\section{Learning Framework and Algorithm}
\label{sec:Learn_Prin_Alg}
In this section, we first present a general GLS correction framework, where a unified view of existing distribution correction frameworks is also provided. Then a kernel-based GLS correction algorithm is proposed to implement the theory-driven learning principle.

\textbf{Learning Framework.}
Based on the main results, we present a theory-driven framework for learning model with GLS correction in this section. As shown in the sufficient condition of GLS correction (i.e., Thm.~\ref{thm:GLS-upper-bound}), an invariant transformation $g$ and an importance weight $w$ are required. Generally, the learning framework for triplet ($g,h,w$), which is equivalent to the upper bound of target error $\varepsilon_Q$, can be formulated as following two principles.
\begin{align}
  &(\textbf{$w$ estimation}) \nonumber \\
  & \mathop{\min }_{w}~ \MC{L}_{\mathrm{Lab}}(w) = D(P^w_Y, Q_Y), ~~ \textrm{s.t.} ~ w \geq 0, ~ \| p^w_Y \|_1 = 1. \label{eq:label_shift} \\
  &(\textbf{GLS correction}) \nonumber \\
  & \mathop{\min }_{h,g}~ \MC{L}_{\mathrm{GLS}}(g,h)= \varepsilon_{P^w}(h\circ g) + \lambda_{g} D(P^w_{Z|Y},Q_{Z|Y}), \label{eq:conditional_shift}
\end{align}
where $\lambda_{g}\geq 0$ is trade-off parameter and $D(\cdot,\cdot)$ is a distance/divergence on distributions. By considering specific parameter settings, the principles above can be connected with other existing learning frameworks, which demonstrates that GLS correction indeed contributes to a more general framework for learning in real-world scenarios.

\textbf{(a)} \textit{Covariate Shift:} by neglecting label shift (i.e., Eq.~\eqref{eq:label_shift} with $w\equiv 1$) and considering the conditional discrepancy as marginal discrepancy, the principles boil down to
\begin{equation}
  \label{eq:covariate_framework}
  \MC{L}_{\mathrm{Mar}}(g,h) = \varepsilon_{P}(h\circ g) + \lambda_{g} D(P_{Z},Q_{Z}),  
\end{equation}
where the model is optimized to be marginal invariant~\cite{ganin2016domain,long2018transferable}, i.e., $g = g_{\mar}$. The sufficiency and necessity can be ensured by~\cite[Thm.~2]{ben2010theory} and Thm.~\ref{thm:zhao_2019_IT-lower-bound}, respectively. 

\textbf{(b)} \textit{Label Shift:} by neglecting conditional shift (i.e., Eq.~\eqref{eq:conditional_shift} with $\lambda_g=0$) and transformation $g$, the principles boil down to 
\begin{equation}
  \label{eq:label_framework}
  \begin{array}{ll}
  &\MC{L}_{\mathrm{Lab}}(w) = D(P^w_Y, Q_Y), ~~ \textrm{s.t.} ~ w \geq 0, ~ \| p^w_Y \|_1 = 1, \\
  &\MC{L}_{\mathrm{Risk}}(h) = \varepsilon_{P^w}(h),  
  \end{array}
\end{equation}
where the model usually learns hypothesis $h$ and estimates weight $w$ alternatively~\cite{lipton2018detecting,yan2017mind}. The sufficiency and necessity can also be ensured by Thm.~\ref{thm:GLS-upper-bound} and~\ref{thm:GLS-lower-bound}, where the conditional shift term is assumed to be zero.

\textbf{(c)} \textit{Conditional Shift:} by neglecting label shift (i.e., Eq.~\eqref{eq:label_shift} with $w\equiv 1$), the principles boil down to 
\begin{equation}
  \label{eq:conditional_framework}
  \MC{L}_{\mathrm{Con}}(g,h) = \varepsilon_{P}(h\circ g) + \lambda_{g} D(P_{Z|Y},Q_{Z|Y}),
\end{equation}
where the representations are required to be conditional invariant to ensure the alignment of local structure~\cite{gong2016domain,luo2021conditional}, i.e., $g=g_{\con}$. Theoretical support can be ensured by Thm.~\ref{thm:GLS-upper-bound} and Thm.~\ref{thm:GLS-lower-bound}, where the label shift term is assumed to be zero.

Several existing works on GLS correction are also essentially connected with Eq.~\eqref{eq:label_shift}-\eqref{eq:conditional_shift}. For different methods, the label shift estimation algorithm BBSE~\cite{lipton2018detecting} is commonly used to estimate $w$. For invariant transformation, Zhang~\etal~\cite{zhang2013domain} and Gong~\etal~\cite{gong2016domain} reduce the conditional discrepancy by reconstructing the marginal distribution with $w$. Tachet des Combes~\etal~\cite{combes2020domain} impose weight $w$ on existing invariant representation learning models. Kirchmeyer~\etal~\cite{kirchmeyer2022mapping} learn $g$ with class-wise OT penalization. Rakotomamonjy~\etal~\cite{rakotomamonjy2021optimal} estimate $w$ with OT-based assignment and learn marginal invariant transformation for $w$-reweighting source. In this work, we propose a kernel-based method KECA for GLS. Compared with methods above, 1) KECA is supported by the derived theorems; 2) the kernel-based conditional embedding theory ensures explicit conditional discrepancy measure and not requires prior assumptions on distributions; 3) KECA explicitly learns the correction model Eq.~\eqref{eq:label_shift}-\eqref{eq:conditional_shift} in a simple but effective way.

{
\begin{algorithm}[!t]
   \caption {Kernel-based GLS Correction Algorithm}\label{alg:GLS_correction}
   \begin{algorithmic}[1]
   \REQUIRE {source data $\{(\MBF{x}_i^s,\MBF{y}_i^s)\}_{i=1}^{{n_s}}$, target data $\{\MBF{x}_i^t\}_{i=1}^{{n_t}}$, maximum iteration $T_{\mathrm{max}}$, learning rate $\lambda$;}
   \ENSURE {invariant transformation $g(\cdot)$, hypothesis $h(\cdot)$, importance weight $\MBF{w}$;}\\
   \STATE Initialize the network parameters $\Theta = \{\Theta_g,\Theta_h\}$; \\
   \FOR {$t=1,2,\ldots,T_{\mathrm{max}}$}
   \STATE \texttt{Forward propagate} $\{\MBF{x}_i^s\}_{i=1}^{{n_s}}$ and $\{\MBF{x}_i^t\}_{i=1}^{{n_t}}$ and obtain $\{(\MBF{z}_i^s, \hat{\MBF{y}}_i^s)\}_{i=1}^{{n_s}}$ and $\{(\MBF{z}_i^t, , \hat{\MBF{y}}_i^t)\}_{i=1}^{{n_t}}$; \\
   {\# \textbf{Fix $g$, $h$. Update $w$.}}
   \STATE \texttt{Compute} the plug-in estimations of $\MBF{q}_{\hat{Y}}$, $\MBF{p}_{\hat{Y}Y}$ and $\MBF{p}_{Y}$ in Eq.~\eqref{eq:BBSE_objective};
   \STATE \texttt{Update} important weight $\MBF{w}$ in Eq.~\eqref{eq:BBSE_objective} via QP or Moore-Penrose inverse; \\
   {\# \textbf{Fix $w$. Update $g$, $h$.}}
   \STATE \texttt{Map} data into RKHS $\MC{H}_{\MC{Z}}\otimes\MC{H}_{\MC{Y}}$ and compute conditional discrepancy as Eq.~\eqref{eq:conditional_metric_alignment}; \\
   \STATE \texttt{Compute} risk objective with conditional discrepancy as $\MC{L}_{\mathrm{GLS}}(g,h)$ in Eq.~\eqref{eq:conditional_shift}; \\
   \STATE \texttt{Update} model via gradient-based optimizer: $\Theta \leftarrow \Theta - \lambda \nabla \MC{L}_{\mathrm{GLS}} (\Theta)$;\\
   \ENDFOR
   \end{algorithmic}
\end{algorithm}
}

\textbf{Algorithm.}
Generally, we consider the DA setting, where a labeled source domain and an unlabeled target domain are accessible during training, and present an algorithm with GLS correction to ensure successful DA. The transformation $g$ and hypothesis $h$ are instantiated as neural networks (NNs) with parameters $\Theta_{g}$ and $\Theta_{h}$, respectively. Since the labels and $Q_Y$ on the target domain are unknown, the pseudo labels assigned as $\hat{Y} = h\circ g(X)$ are usually employed. Besides, the weight estimation in Eq.~\eqref{eq:label_shift} is usually implemented via the predictor $h\circ g$. Then, the optimization of Eq.~\eqref{eq:label_shift}-\eqref{eq:conditional_shift} is divided into two alternative learning problems as Alg.~\ref{alg:GLS_correction}.

\textbf{(a)} \textbf{$w$ estimation.}
This stage can be taken as update of $w$ with fixed $g\circ h$, where the objective Eq.~\eqref{eq:label_shift} is considered. This problem can be efficiently solved by the BBSE~\cite{lipton2018detecting} algorithm which also admits some nice properties, e.g., consistency of estimation and detection of label shift. Moreover, the property of detecting label shift~\cite[Prop.~4]{lipton2018detecting} ensures the objective of BBSE is actually a measure between $P^w_Y$ and $Q_Y$, e.g., it is equivalent to Eq.~\eqref{eq:label_shift}. Assuming $K=|\MC{Y}|$ is finite, the PDFs of $Y$ and $\hat{Y}$ boil down to stochastic vectors, e.g., $\MBF{p}_{Y},\MBF{p}_{\hat{Y}} \in \MBB{R}_{+}^{K}$. Then, BBSE considers the joint distribution over $Y$ and $\hat{Y}=h(Z)$ as $\MBF{p}_{\hat{Y}Y}\in \MBB{R}^{K\times K}$ and provide a specific form of Eq.~\eqref{eq:label_shift} as
\begin{equation}
  \label{eq:BBSE_objective}
   \mathop{\min }_{\MBF{w}} ~\| \MBF{q}_{\hat{Y}} - \MBF{p}_{\hat{Y}Y} \MBF{w} \|_2^2, \quad \textrm{s.t.} \quad  \MBF{w} \in \MBB{R}^K_{+}, ~~ \MBF{w}^T \MBF{p}_{Y} = 1.
\end{equation}
Eq.~\eqref{eq:BBSE_objective} can be efficiently solved by Quadratic-Programming (QP)~\cite{combes2020domain}. For empirical estimation, $\MBF{q}_{\hat{Y}}$, $\MBF{p}_{\hat{Y}Y}$ and $\MBF{p}_{Y}$ are computed via plug-in estimation, where the convergence has been proved by Lipton~\etal~\cite{lipton2018detecting}.

\textbf{(b)} \textbf{GLS correction.}
This stage can be taken as update of transformation and hypothesis, i.e., $g\circ h$, with fixed $w$, where the objective Eq.~\eqref{eq:conditional_shift} is considered. To ensure the efficiency of discrepancy estimation and optimization, we employ the conditional embedding statistics in Reproducing Kernel Hilbert Space (RKHS)~\cite{muandet2017kernel,park2020measure,luo2021conditional,klebanov2020rigorous} to characterize the conditional distributions $P^w_{Z|Y}$ and $Q_{Z|Y}$. Then the conditional discrepancy $D(\cdot,\cdot)$ is measured by the CMMD metric $d_{\mathrm{CMMD}}(\cdot,\cdot)$~\cite{ren2016conditional,ren2021learning,xu2021cross} between the first order statistic, i.e.,
\begin{align}
  D(P^w_{Z|Y},Q_{Z|Y}) &= d^2_{\mathrm{CMMD}}(P^w_{Z|Y},Q_{Z|Y}) \nonumber \\
  &= \| \MC{U}_{P^w_{Z|Y}} -  \MC{U}_{Q_{Z|Y}} \|^2_{\MC{H}_{\MC{Z}}\otimes\MC{H}_{\MC{Y}}}, \label{eq:conditional_metric_alignment}
\end{align}
where $\MC{H}_{\MC{Z}}$ and $\MC{H}_{\MC{Y}}$ are the RKHSs induced by kernels $k_\MC{Z}$ and $k_\MC{Y}$; $\MC{U}$ is the embedding operator of conditional mean in RKHS. Since there are no labels on the target domain, $Q_{Z|Y}$ is estimated with the pseudo labels $\hat{Y}$, which is proved to be effective in empirical applications \cite{long2018conditional,combes2020domain}. More details about the kernel metrics and empirical estimations are provided in supplemental material. The main advantages of kernel metrics are analytic expression for computation and smoothness for optimization with proper kernels, e.g., Gaussian kernel.

\begin{figure*}[t]
  \centering
  \subfloat[Shift Comparison \label{fig:shift_abla_curve}]{\includegraphics[width=0.245\linewidth,trim=85 40 50 70,clip]{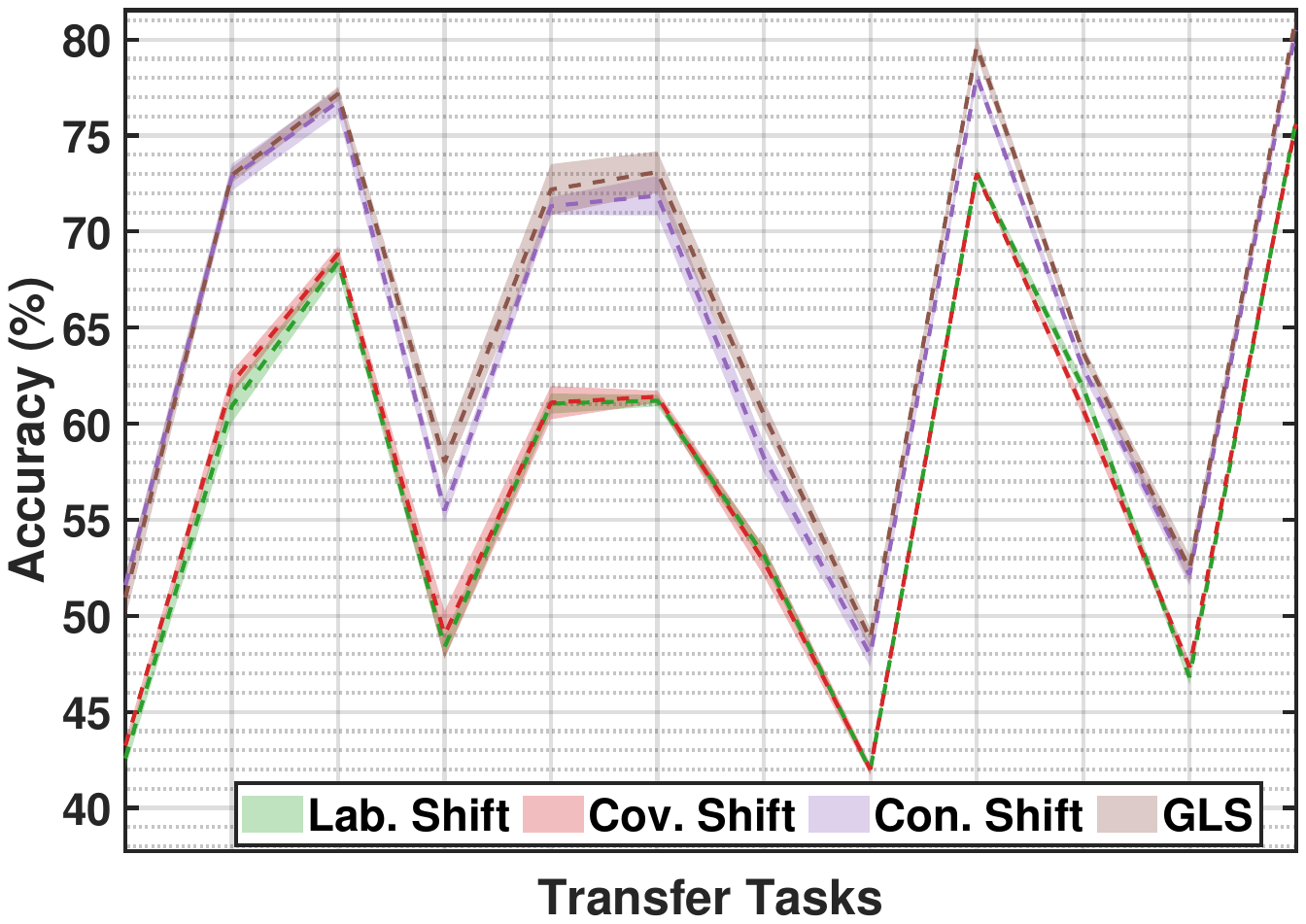}}
  \hfill
  \subfloat[Shift Comparison \label{fig:shift_abla_bar}]{\includegraphics[width=0.245\linewidth,trim=85 40 50 70,clip]{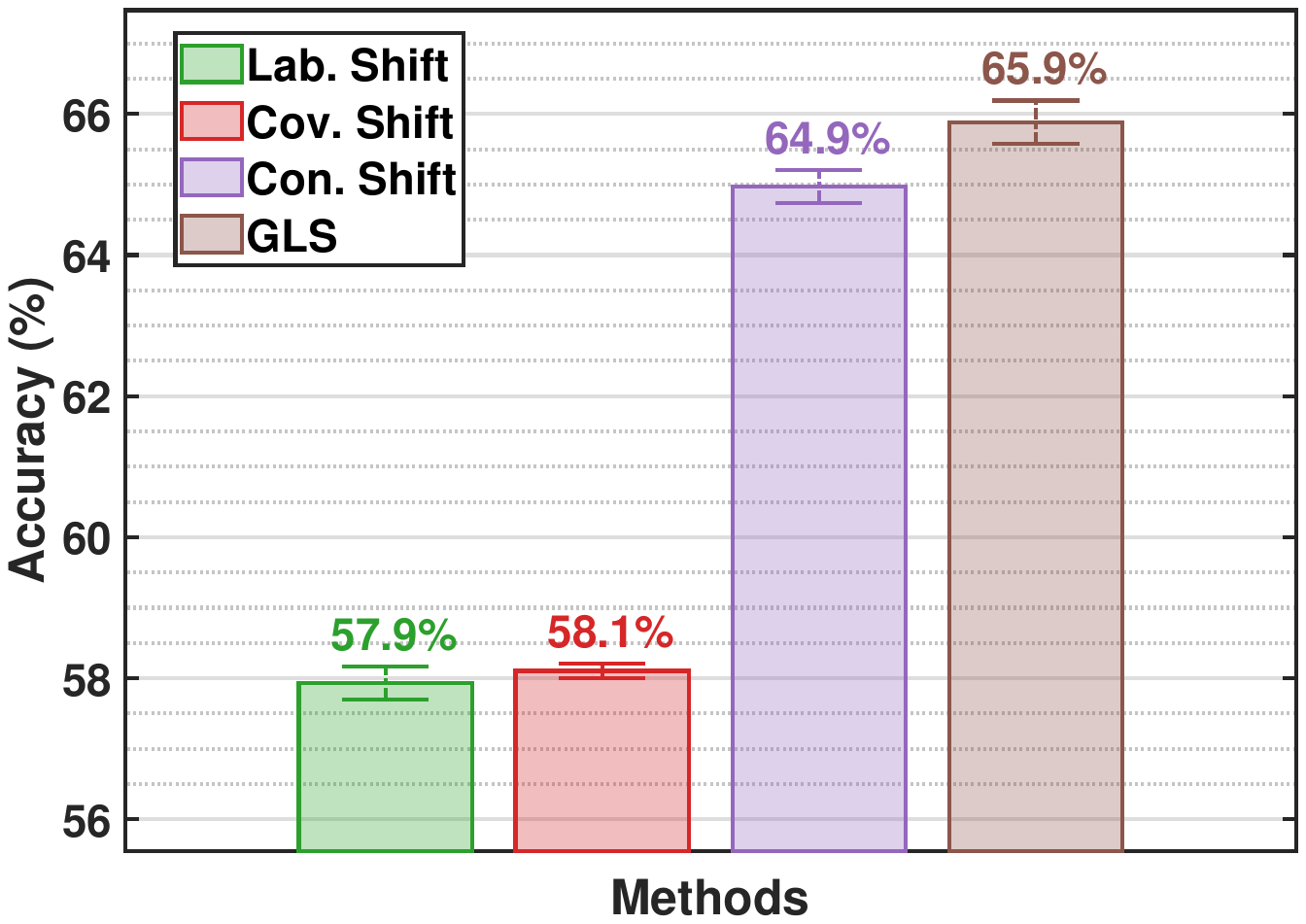}}
  \hfill
  \subfloat[Subsampled Source \label{fig:label_shift_exp_subsource}]{\includegraphics[width=0.245\linewidth,trim=85 40 50 70,clip]{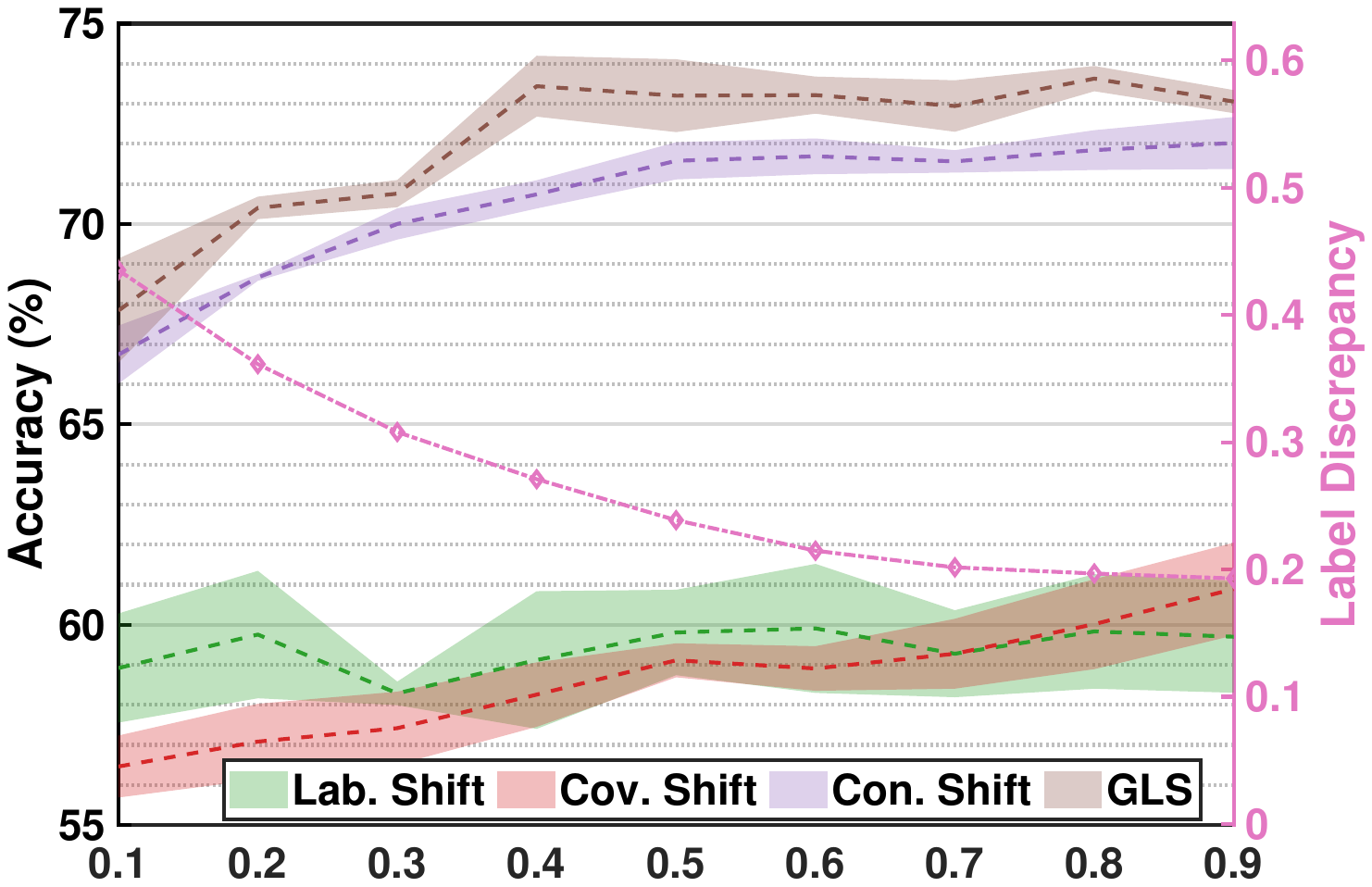}}
  \hfill
  \subfloat[Subsampled Target \label{fig:label_shift_exp_subtarget}]{\includegraphics[width=0.245\linewidth,trim=85 40 50 70,clip]{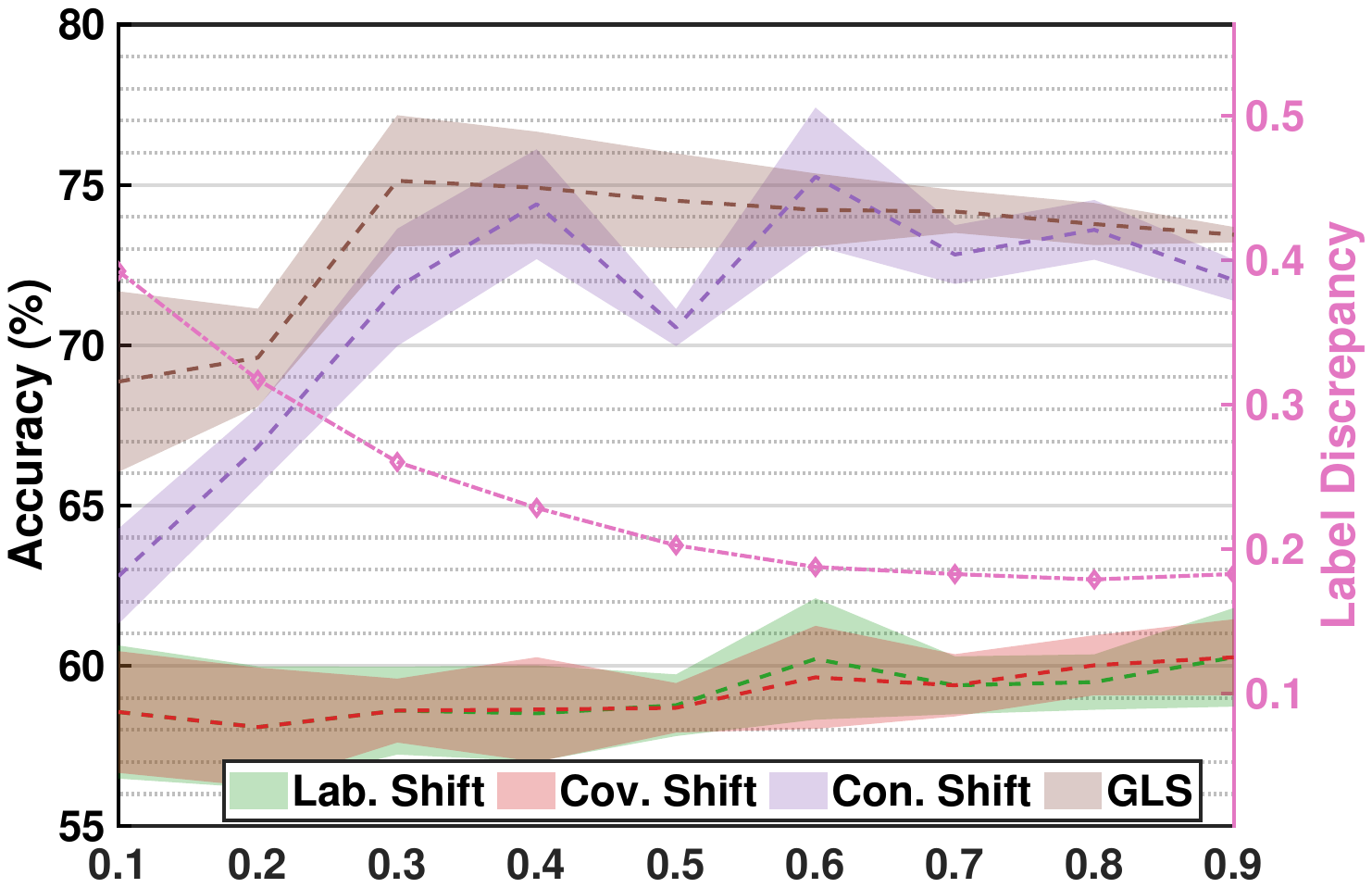}}
  \\
  \centering
  \subfloat[w/o label reweighting \label{fig:abla_DB_wo_weight}]{\includegraphics[height=85pt,width=0.245\linewidth,trim=55 35 40 30,clip]{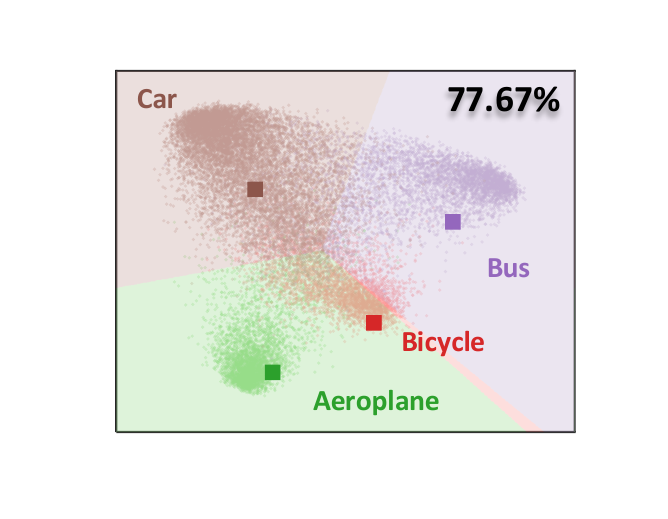}}
  \hfill
  \subfloat[w/o conditional matching \label{fig:abla_DB_wo_match}]{\includegraphics[height=85pt,width=0.245\linewidth,trim=55 35 40 30,clip]{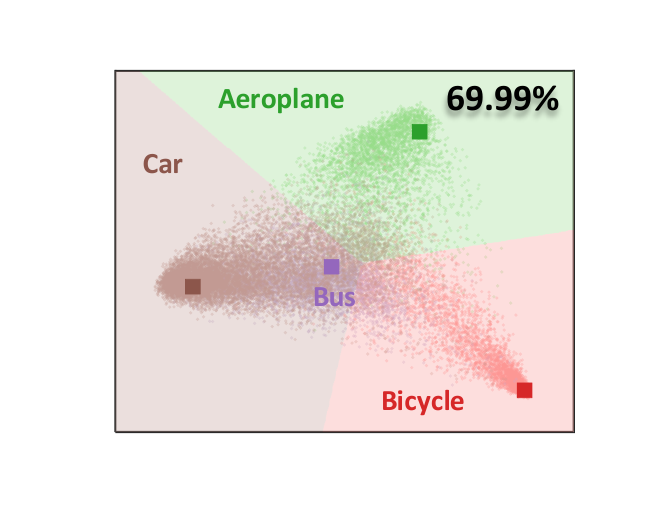}}
  \hfill
  \subfloat[KECA \label{fig:abla_DB_GLS}]{\includegraphics[height=85pt,width=0.245\linewidth,trim=55 35 40 30,clip]{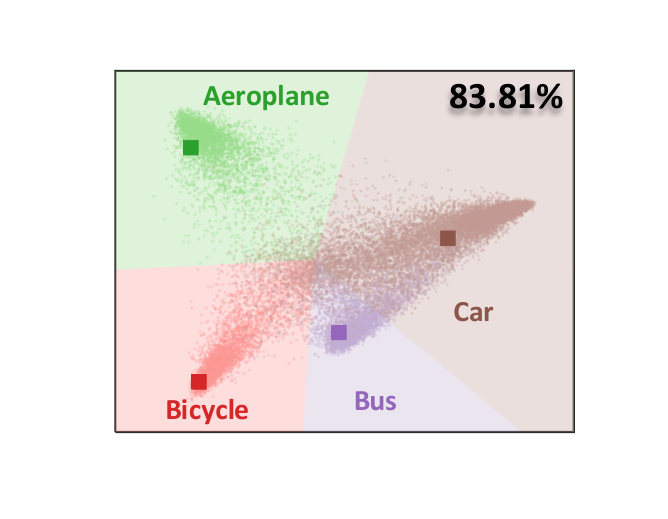}}
  \hfill
  \subfloat[KECA w/ $w^*$ \label{fig:abla_DB_oracle}]{\includegraphics[height=85pt,width=0.245\linewidth,trim=55 35 40 30,clip]{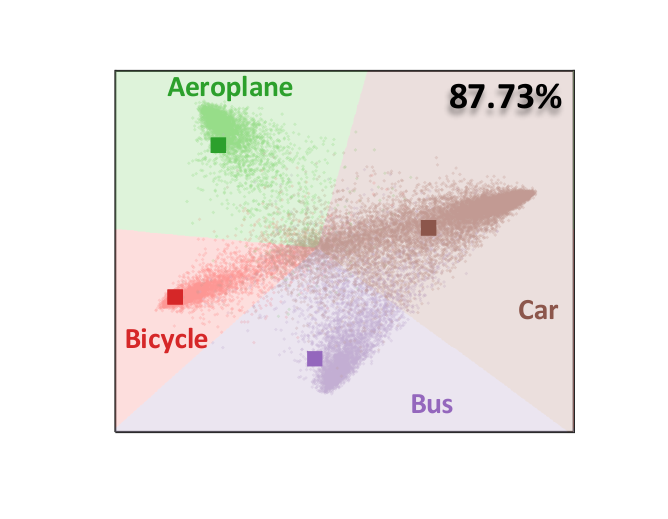}}
  \\
  \caption{(a)-(b): Accuracy curves/bars and 95\% confidence intervals of different dataset shift correction models. (c)-(d): Accuracy curves and label discrepancies $d_{\TV}(P_Y,Q_Y)$ with different subsampled rates and domains. (e)-(h): Visualization of representations and decision boundaries. `$\blacksquare$': class-wise means on source domain, `$\bullet$': target samples, `background color': decision boundary.}
  \label{fig:abla_DB&shift&label_exp}
\end{figure*}

\section{Numerical Experiment}
\label{sec:experiment}
In this section, we conduct numerical evaluations on standard GLS datasets. We first present the implementation details in Sec.~\ref{subsec:set_up}. Then the evaluation experiments on theoretical results, analysis of different shift correction frameworks, and the comparison experiments with SOTA GSL correction models are presented in Sec.~\ref{subsec:empirical_analysis}.

\subsection{Set Up}
\label{subsec:set_up}

\textbf{Datasets.}
We conducted experiments on standard visual GLS datasets, i.e., Office-31~\cite{saenko2010adapting}, Office-Home~\cite{venkateswara2017deep} and VisDA-2017~\cite{peng2017visda}. Following common protocols~\cite{combes2020domain,rakotomamonjy2021optimal,kirchmeyer2022mapping}: 1) ResNet-50~\cite{he2016deep} is used as backbone network for feature extraction; 2) to simulate the GLS scenarios, the subsampled source domain consists of 30\% data of the first $K_1$ classes (alphabetical order) and all data in remaining $K_2$ classes. We use the random seeds released by literature~\cite{combes2020domain,rakotomamonjy2021optimal,kirchmeyer2022mapping} to generate the same GLS scenarios. 

\begin{itemize}
  \item \textbf{Office-Home}~consists of about 15k images from 4 domains with 65 classes, i.e., Artistic (\textbf{Ar}), Clipart (\textbf{Cl}), Product (\textbf{Pr}) and Real-World (\textbf{Rw}). The subsampled source domains with parameter pair $(K_1,K_2)$=$(32,33)$ are denoted as \textbf{sAr}, \textbf{sCl}, \textbf{sPr} and \textbf{sRw}. 

  \item \textbf{Office-31}~consists of about 4k images from 3 domains with 31 classes, i.e., Amazon (\textbf{A}), Web camera (\textbf{W}) and Digital SLR camera (\textbf{D}). The subsampled source domains with $(K_1,K_2)$=$(15,16)$ are denoted as \textbf{sA}, \textbf{sD} and \textbf{sW}.
  
  \item \textbf{VisDA-2017}~consists of about 152k synthetic images from source domain Synthetic (\textbf{S}) and 55k real-world images from target domain Real (\textbf{R}). The subsampled source domain with $(K_1,K_2)$=$(6,6)$ is denoted as \textbf{sS}. 
\end{itemize}

\textbf{Network Architectures.}
We implement the NN-based learning algorithm in PyTorch platform \cite{paszke2019pytorch}. The transformation $g$ consists of a deep neural network (i.e., ResNet-50) and two Fully-Connected (FC) layers. The first FC layer project representations from $\MBB{R}^{2048}$ to $\MBB{R}^{1024}$ with batch normalization and Leaky ReLU activation ($\alpha = 0.2$); the second FC layer from $\MBB{R}^{1024}$ to $\MBB{R}^{512}$ with batch normalization and Tanh activation. The hypothesis is an FC layer which projects representation from $\MBB{R}^{512}$ to probability prediction $\MBB{R}^{|\MC{Y}|}$ with softmax activation. All experiments are implemented on an Ubuntu 18.04 operating system PC with an Intel Core i7-6950X 3.00GHz CPU, 64GB RAM and an NVIDIA TITAN Xp GPU.

\textbf{Algorithm. }
Gaussian kernel $k(\MBF{z}_1,\MBF{z}_2) = \exp (-\| \MBF{z}_1 - \MBF{z}_2 \|_2^2/\sigma)$ is employed to ensure the metric property and smoothness of conditional discrepancy in Eq.~\eqref{eq:conditional_metric_alignment}. The networks are optimized via gradient descent with ADAM optimizer. The learning rate $\lambda$ in Alg.~\ref{alg:GLS_correction} and conditional matching parameter $\lambda_g$ in Eq.~\eqref{eq:conditional_shift} are set as 8$e$-4 and 1$e$-1, respectively. To reduce the uncertainty of pseudo label in GLS correction, we first pretrain the model $(g,h)$ with the empirical risk on the source domain for warm-up, where the warm-up epochs are set as 40 empirically. Then the pretrained parameters $\{\Theta_g,\Theta_h\}$ are applied as the initialization in Alg.~\ref{alg:GLS_correction}. Following the standard protocols for GLS~\cite{combes2020domain,rakotomamonjy2021optimal,kirchmeyer2022mapping}, all experiments are randomly repeated for 10 times and the mean results are reported.

\begin{table*}[t]
  \caption{Comparison of different subsampled rates. Classification accuracies and standard errors (\%) and TV distances ($d_{\TV}\in[0,1]$) between label distributions on submsampled Office-Home dataset (ResNet-50).}
  \label{tab:subsampled_rate}
  \centering
  \renewcommand{\tabcolsep}{0.3pc}
  \begin{tabular}{cl|cccccccccc}
    \toprule
    \rowcolor{gray!25}
    & \textbf{Subsampled Rates} & 10\% & 20\% & 30\% & 40\% & 50\% & 60\% & 70\% & 80\% & 90\% & Avg. \\
    \multirowcell{5}{\textbf{Subsampled} \\ \textbf{Source Domain}} & $d_{\TV}(P_Y,Q_Y)$ & 0.44 & 0.36 & 0.31 & 0.27 & 0.24 & 0.21 & 0.20 & 0.20 & 0.19 & 0.27 \\
    & Lab. Shift & 56.8$\pm$0.7 & 57.6$\pm$0.8 & 57.7$\pm$0.2 & 57.0$\pm$0.9 & 58.1$\pm$0.5 & 57.8$\pm$0.8 & 57.6$\pm$0.6 & 57.3$\pm$0.7 & 57.8$\pm$0.7 & 57.5 \\
    & Cov. Shift & 55.3$\pm$0.4 & 55.9$\pm$0.5 & 56.4$\pm$0.5 & 57.3$\pm$0.4 & 58.5$\pm$0.2 & 58.0$\pm$0.3 & 58.3$\pm$0.4 & 58.6$\pm$0.6 & 58.6$\pm$0.6 & 57.4 \\
    & Con. Shift & 65.4$\pm$0.4 & 68.5$\pm$0.0 & 69.3$\pm$0.2 & 70.3$\pm$0.2 & 70.6$\pm$0.2 & 71.1$\pm$0.2 & 71.1$\pm$0.1 & 71.3$\pm$0.3 & 71.4$\pm$0.3 & 69.9 \\
    & GLS & \textbf{66.5}$\pm$0.7 & \textbf{69.9}$\pm$0.1 & \textbf{70.3}$\pm$0.2 & \textbf{72.5}$\pm$0.4 & \textbf{72.5}$\pm$0.5 & \textbf{72.8}$\pm$0.2 & \textbf{72.5}$\pm$0.3 & \textbf{73.0}$\pm$0.2 & \textbf{72.8}$\pm$0.1 & \textbf{71.4} \\
    \midrule
    \rowcolor{gray!25}
    & \textbf{Subsampled Rates} & 10\% & 20\% & 30\% & 40\% & 50\% & 60\% & 70\% & 80\% & 90\% & Avg. \\
    \multirowcell{5}{\textbf{Subsampled} \\ \textbf{Target Domain}} & $d_{\TV}(P_Y,Q_Y)$ & 0.39 & 0.32 & 0.26 & 0.23 & 0.20 & 0.19 & 0.18 & 0.18 & 0.18 & 0.24 \\
    & Lab. Shift & 55.4$\pm$1.1 & 55.7$\pm$1.0 & 56.5$\pm$0.7 & 56.3$\pm$0.8 & 57.6$\pm$0.5 & 57.3$\pm$1.0 & 57.8$\pm$0.5 & 57.9$\pm$0.4 & 57.5$\pm$0.8 & 56.9 \\
    & Cov. Shift & 55.9$\pm$1.0 & 55.7$\pm$0.9 & 57.1$\pm$0.5 & 56.9$\pm$0.8 & 57.3$\pm$0.4 & 57.5$\pm$0.8 & 58.3$\pm$0.5 & 58.9$\pm$0.5 & 58.6$\pm$0.6 & 57.4 \\
    & Con. Shift & 60.6$\pm$0.8 & 64.8$\pm$0.6 & 68.3$\pm$0.9 & 72.0$\pm$0.9 & 69.9$\pm$0.3 & 72.6$\pm$1.1 & 71.3$\pm$0.5 & 72.4$\pm$0.5 & 71.1$\pm$0.3 & 69.2 \\
    & GLS & \textbf{65.4}$\pm$1.4 & \textbf{68.0}$\pm$0.8 & \textbf{71.4}$\pm$1.0 & \textbf{72.8}$\pm$0.9 & \textbf{72.6}$\pm$0.8 & \textbf{73.2}$\pm$0.6 & \textbf{73.3}$\pm$0.3 & \textbf{73.3}$\pm$0.3 & \textbf{73.0}$\pm$0.1 & \textbf{71.4} \\
    \bottomrule
  \end{tabular}
\end{table*}

\subsection{Empirical Evaluation and Analysis}
\label{subsec:empirical_analysis} 

In this section, we conduct experiments to validate the theoretical results and GLS framework from several aspects.

\textbf{Shift Assumptions.}
To explicitly compare the different shift correction frameworks discussed in Sec.~\ref{sec:Learn_Prin_Alg}, we present the results of models based on label shift Eq.~\eqref{eq:label_framework}, covariate shift Eq.~\eqref{eq:covariate_framework}, conditional shift Eq.~\eqref{eq:conditional_framework} and GLS. The results in Fig.~\ref{fig:shift_abla_curve}-\ref{fig:shift_abla_bar} show the accuracy curves and mean accuracy bars over 12 transfer tasks on Office-Home. Since label shift and covariate shift correction models consider the marginal distribution shifts on $Y$ and $X$, they achieve similar results under GLS scenario. However, since marginal invariant learning frameworks usually requires strong assumption on conditional distribution, they are significantly less practical for the dataset shift in real-world scenarios. Hopefully, the conditional shift correction model focuses on the local structure alignment, which is effective in alleviating negative transfer induced by marginal alignment. Thus, it can be observed that conditional shift correction significantly improves the mean task accuracy to $64.9\%$, which demonstrates that the transfer of semantic information is important and necessary (i.e., Thm.~\ref{thm:GLS-lower-bound}). Further, the GLS correction model mitigates the biased estimation of target risk and improves the accuracy to $65.9\%$, which implies that conditional shift correction is not optimal and indeed insufficient. It also demonstrates that GLS correction usually ensures lower generalization error compared other shift correction frameworks, i.e., the upper-bound and sufficient theorem Thm.~\ref{thm:GLS-upper-bound} are valid in empirical scenarios. Overall, these results validate the main results on limitations of invariant representation learning and the sufficiency/necessity of GLS correction for successful knowledge transfer.

\textbf{Impacts of Label Shift.}
To study the impacts of label shift on invariant representation learning, we compare the shift correction frameworks under different degrees of label shift (i.e., changing the subsampled rates and subsampled domains) on VisDA-2017 sS$\to$R. The results on subsampled source domain and subsampled target domain are presented in Tab.~\ref{tab:subsampled_rate} and Fig.~\ref{fig:label_shift_exp_subsource}-\ref{fig:label_shift_exp_subtarget}, where the label distribution discrepancy curves $d_{\TV}(P_Y,Q_Y)$ under different subsampled rates (i.e., pink dashed curves) are provided. We observe that when the subsampled rates are small, i.e., the label shift is severe, the label shift model is superior to the covariate shift model, and the GLS model is significantly better than the conditional shift model. These observations demonstrate that: 1) invariant transformations $g_{\mar}$ and $g_{\con}$ are unreliable when label shift exists as Prop.~\ref{prop:imposs-joint-matching}; 2) the generalization performance is actually decided by the label shift error and conditional shift error as Thm.~\ref{thm:GLS-upper-bound}, then the conditional shift framework without label correction cannot address the error induced by $d_{\TV}(P_Y,Q_Y)$. Moreover, for mean results over all subsampled rates, the models with label shift correction, i.e., weighting strategy, generally outperform the invariant transformation models without operations on label $Y$. These results demonstrate that label shift does have non-negligible effects on invariant representation learning.  

\textbf{Decision Boundaries.}
To empirically analyze the effects of different modules in GLS correction, we conduct ablation experiment to visualize the decision boundaries learned by different models, i.e., KECA w/o weighting, KECA w/o conditional matching, KECA, and KECA w/ $w^*$ (i.e., oracle weight) on VisDA-2017 sS$\to$R, where four classes are selected. The results in Fig.~\ref{fig:abla_DB_wo_weight}-\ref{fig:abla_DB_oracle} show that the model w/o conditional matching may induce the misalignment problem, i.e., the samples of `\textbf{Bus}' are misaligned to `\textbf{Bicycle}' and `\textbf{Car}' with similar semantic information in Fig.~\ref{fig:abla_DB_wo_match}, which demonstrates that $g_{\con}$ is crucial for identifying and preserving the local structures during transfer. Besides, it can be observed from Fig.~\ref{fig:abla_DB_GLS}-\ref{fig:abla_DB_oracle} that KECA with estimated weight actually learns a similar decision boundary with model with oracle label weight $w^*$. It demonstrates that GLS correction with $w$ further encourages model to learn a proper decision boundary for the target distribution compared with other learning frameworks, i.e., conditional shift correction. Finally, the model with oracle weight $w^*$ ensures more discriminative representations with a consistent target risk estimation as Fig.~\ref{fig:abla_DB_oracle}. These qualitative results demonstrate that 1) the two main components for GLS, i.e., matching and reweighting, are both crucial for the successful transfer; 2) the two components indeed ensure different empirical effects in representation space and hypothesis space, i.e., better estimation of cluster correlations and decision risks.

\begin{figure*}[t]
  \centering
  \subfloat[KECA (sA$\to$D) \label{fig:feavis_31_AD_KECA}]{\includegraphics[height=95pt,width=0.24\linewidth,trim=140 290 125 280,clip]{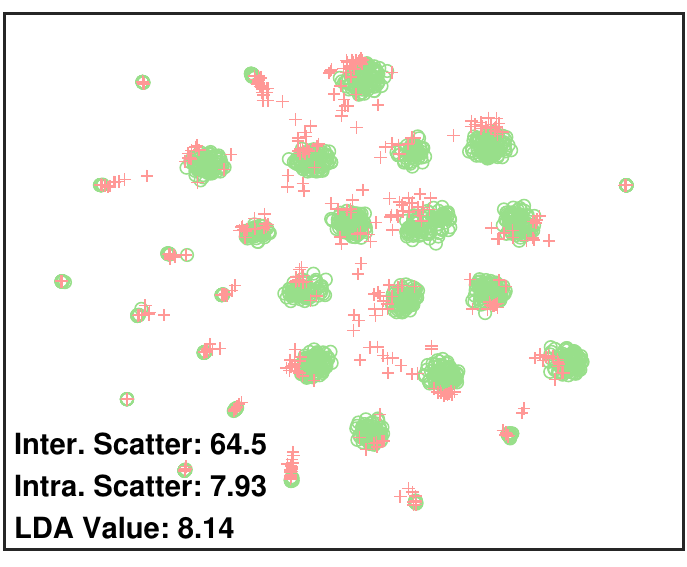}}
  \hfill
  \subfloat[Oracle (sA$\to$D) \label{fig:feavis_31_AD_oracle}]{\includegraphics[height=95pt,width=0.24\linewidth,trim=140 290 125 280,clip]{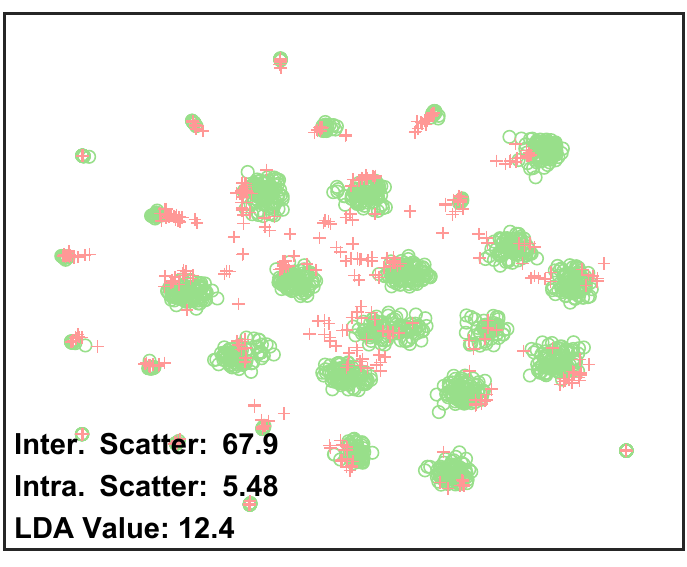}}
  \hfill
  \subfloat[KECA (sA$\to$W) \label{fig:feavis_31_AW_oracle}]{\includegraphics[height=95pt,width=0.24\linewidth,trim=140 290 125 280,clip]{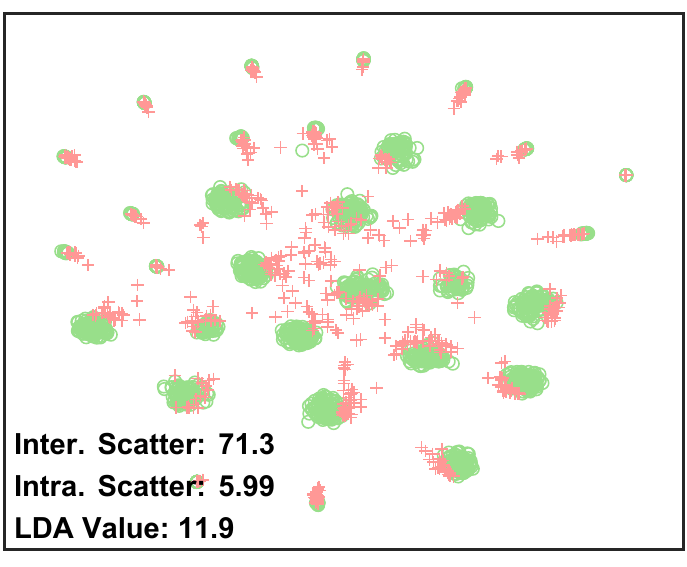}}
  \hfill
  \subfloat[Oracle (sA$\to$W) \label{fig:feavis_31_AW_oracle}]{\includegraphics[height=95pt,width=0.24\linewidth,trim=140 290 125 280,clip]{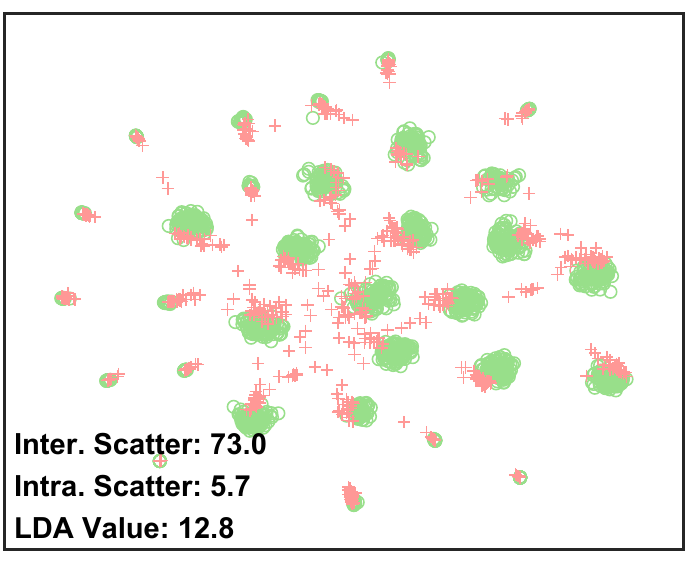}}
  \\
  \centering
  \subfloat[Discrepancy sA$\to$D \label{fig:DivCurve_31_AD}]{\includegraphics[width=0.245\linewidth,trim=85 40 50 70,clip]{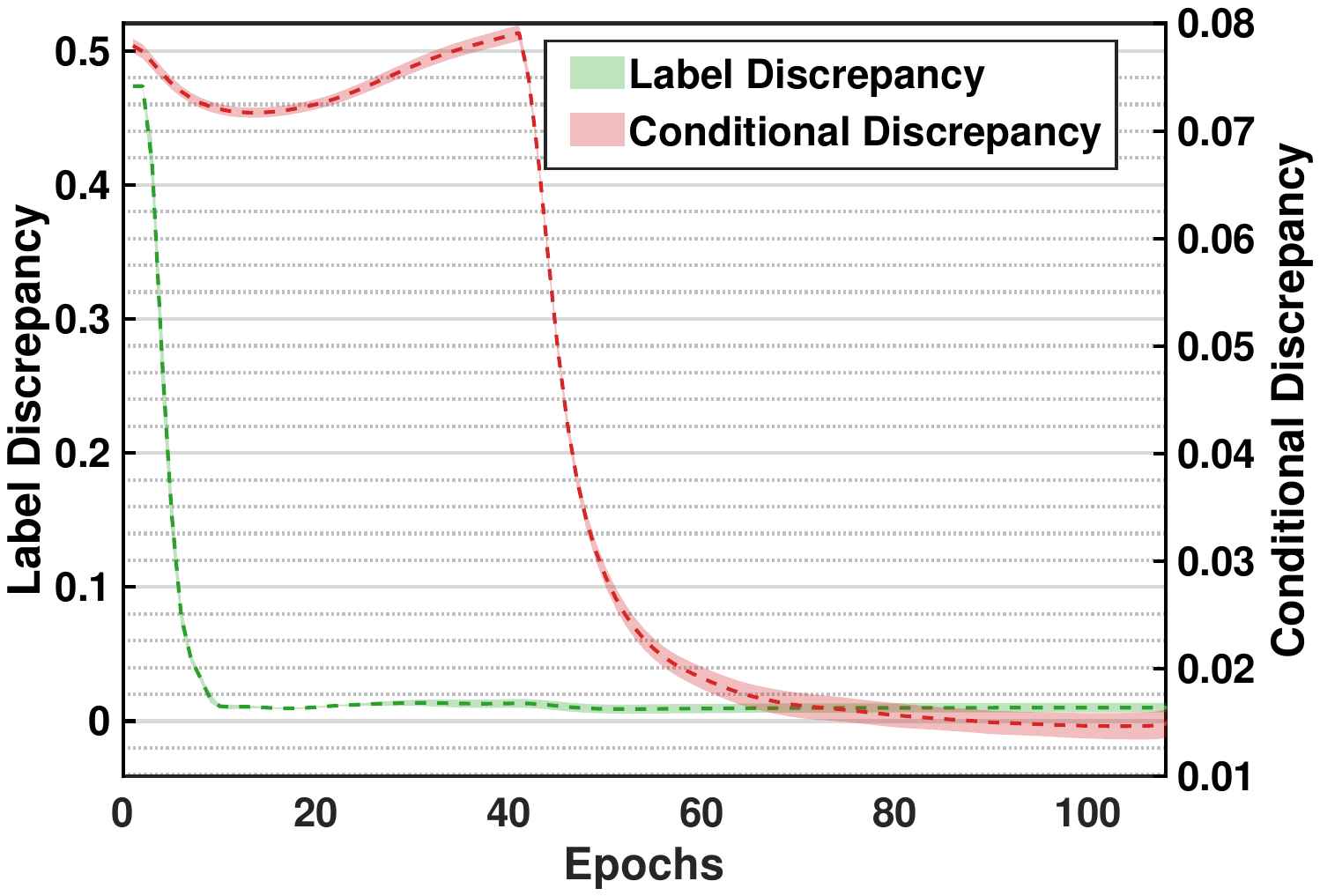}}
  \hfill
  \subfloat[Discrepancy sD$\to$A \label{fig:DivCurve_31_DA}]{\includegraphics[width=0.245\linewidth,trim=65 40 50 70,clip]{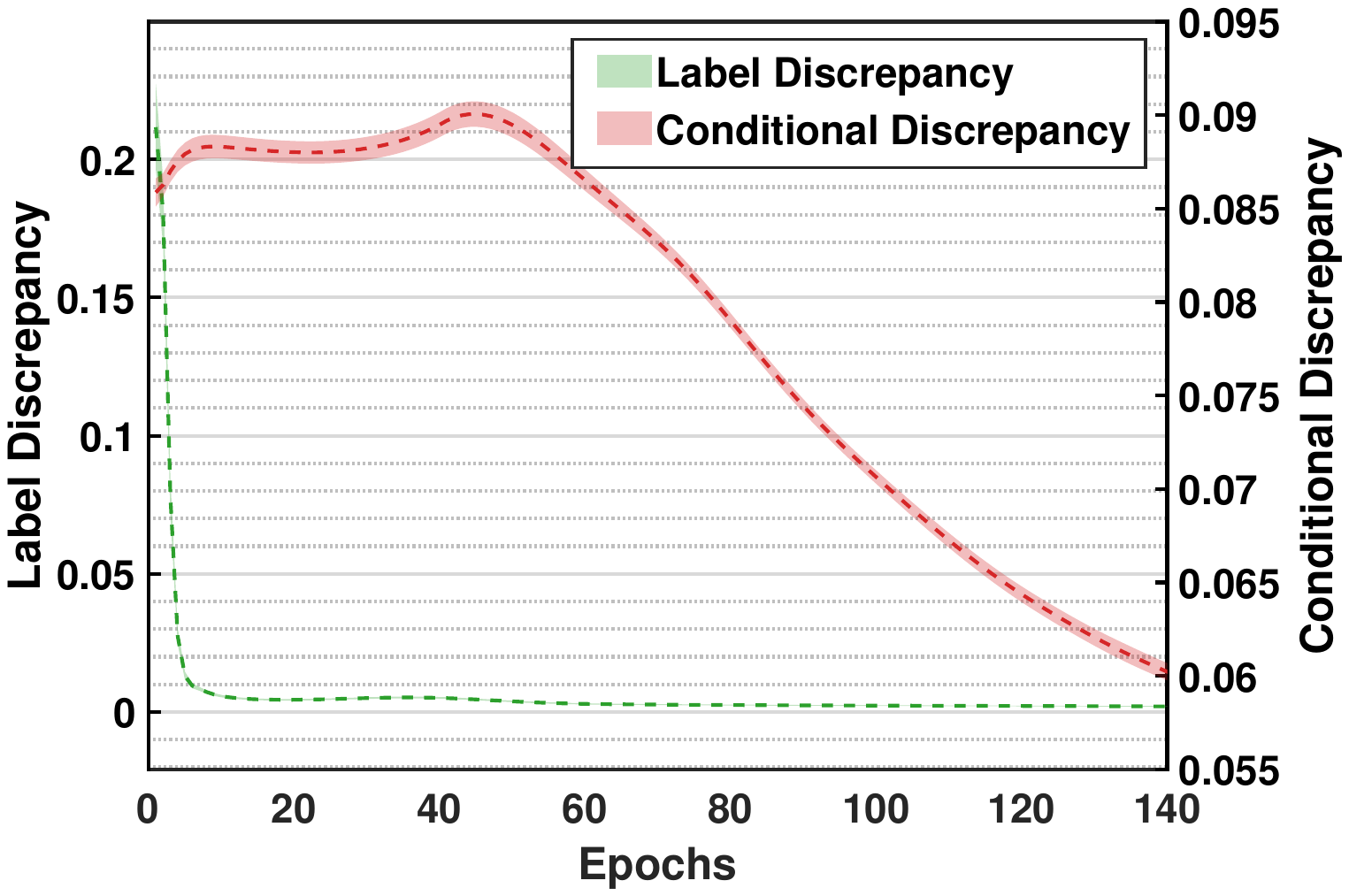}}
  \hfill
  \subfloat[Discrepancy sA$\to$D \label{fig:DivCurve_31_AD_AfterWarmup}]{\includegraphics[width=0.245\linewidth,trim=55 40 50 70,clip]{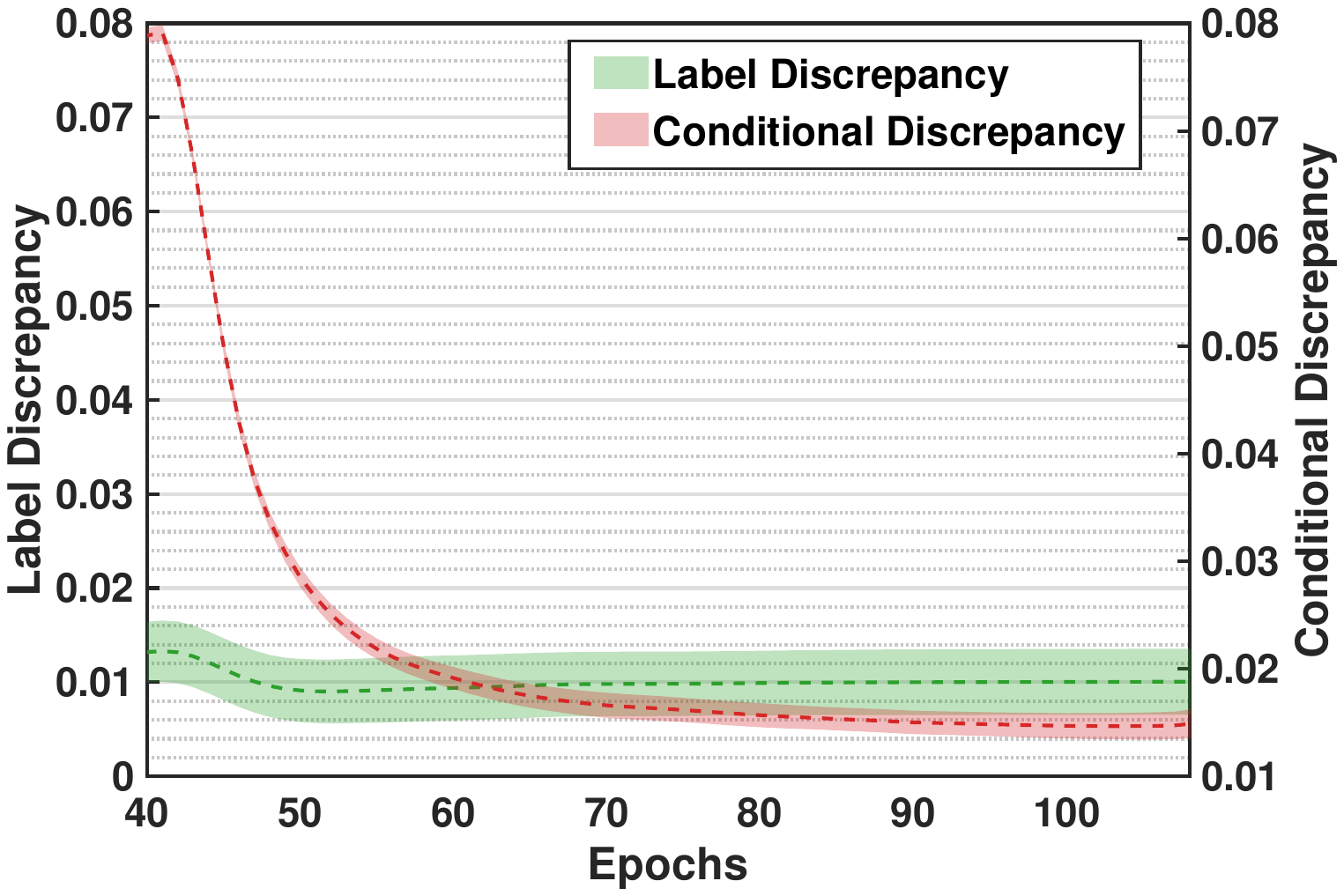}}
  \hfill
  \subfloat[Discrepancy sD$\to$A \label{fig:DivCurve_31_DA_AfterWarmup}]{\includegraphics[width=0.245\linewidth,trim=85 40 50 50,clip]{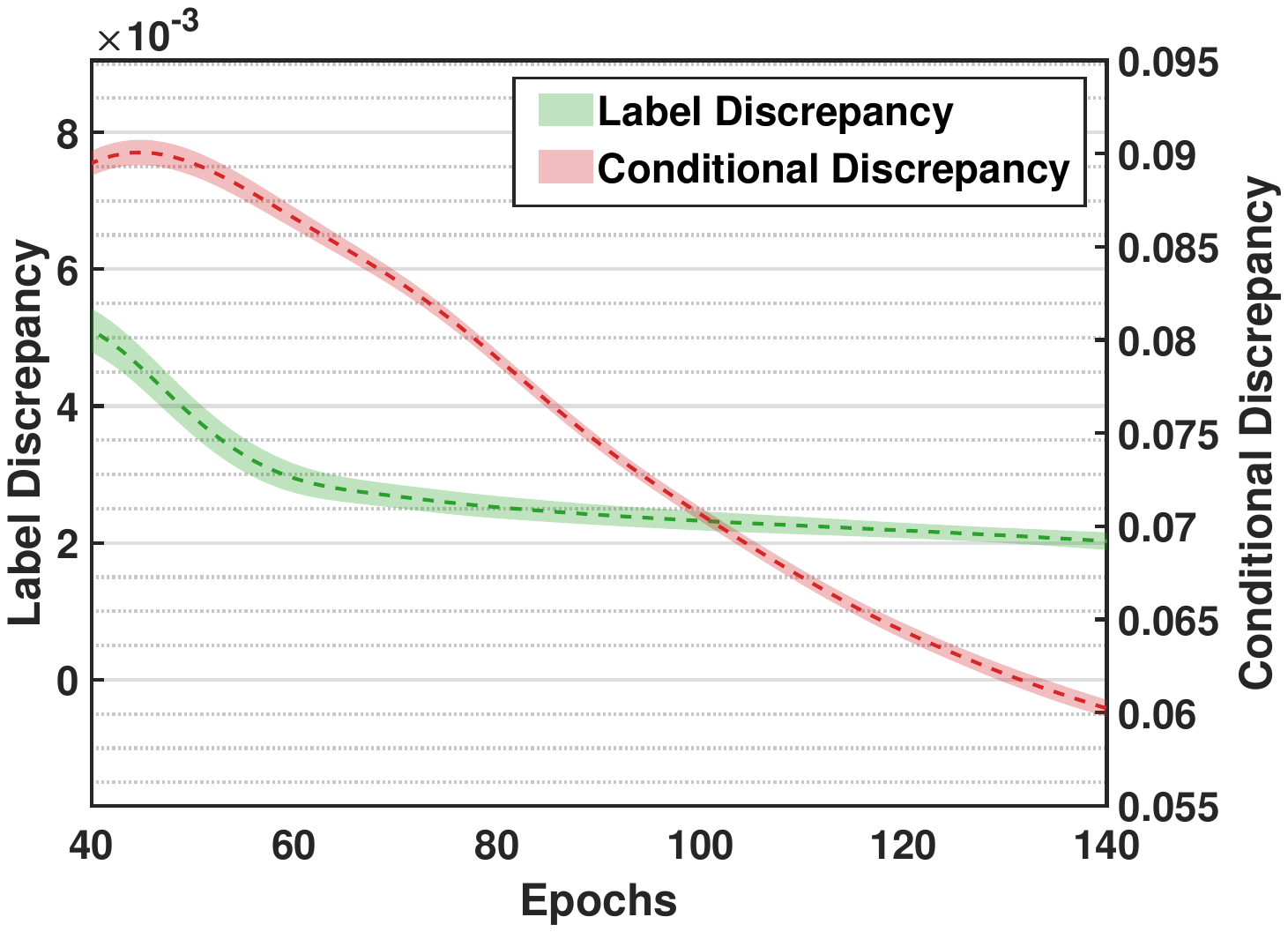}}
  \\
  \caption{(a)-(d): Visualization of representations of KECA model and Orcale model (i.e., KECA with ground-truth weight $w^*$) via t-SNE dimensionality reduction algorithm~\cite{van2008visualizing} on Office-31. \textcolor{green}{`$\circ$'}: source samples; \textcolor{red}{`$+$'}: target samples. (e)-(h): Curves of label discrepancy $d_{\JS,a}(P^w_Y,Q_Y)$ and conditional discrepancy $D(P^w_{Z|Y},Q_{Z|Y})$ on Office-31 dataset. (e)-(f): curves of complete training process; (g)-(h): curves after the warm-up stage where the conditional matching and importance weight are applied.}
  \label{fig:fea_visualization}
\end{figure*}

\textbf{{Selection of Kernel Functions.}}
Though kernel-based embedding theory and discrepancy measures have appealing theoretical properties, it is still necessary to evaluate the empirical performance of KECA under different selections of kernel functions $k(\cdot,\cdot)$. Thus, we conduct experiments on Office-31 to compare several commonly used kernel function including linear kernel, polynomial kernel of degree 2, Laplacian kernel and Gaussian Kernel. The results are presented in Tab.~\ref{tab:selection_kernel_functions}. Form theoretical aspects, universal property of kernel function $k$ is usually required to ensure the distribution metric property for kernel-based metric (e.g., MMD~\cite{gretton2012kernel} and CMMD~\cite{ren2016conditional}). Among these kernels, Laplacian kernel and Gaussian kernel are better since they are universal kernels, which ensure the distribution embedding property for rigorous discrepancy measure. From empirical aspects, the results show that polynomial kernel and Gaussian kernel are better than other kernels, where Gaussian kernel outperforms others on 5 of 6 transfer tasks. In terms of optimization, Gaussian kernel and polynomial kernel are better due to the smoothness of the function. In conclusion, Gaussian kernel is generally better than other kernel functions in the views of theoretical guarantee and empirical performance.

\begin{table}
  \centering
  \caption{Classification accuracies (\%) of KECA model with different kernel functions on Office-31 dataset (ResNet-50).}
  \label{tab:selection_kernel_functions}
  \renewcommand{\tabcolsep}{0.15pc} 
  \begin{tabular}{l|ccccccc}
    \toprule
    \rowcolor{gray!25}
    \textbf{Tasks} & sA$\to$D & sA$\to$W & sD$\to$A & sD$\to$W & sW$\to$A & sW$\to$D & Avg. \\
    \rowcolor{gray!25}
    $d_{\TV}(P_Y,Q_Y)$ & 0.29 & 0.26 & 0.28 & 0.29 & 0.28 & 0.30 & 0.28 \\
    Linear & 85.1 & 89.5 & 70.5 & 96.8 & 71.0 & 98.0 & 85.2 \\
    Polynomial & 86.0 & \textbf{89.8} & 74.4 & 97.5 & 74.3 & 98.0 & \textbf{86.7} \\
    Laplacian & 83.8 & 87.9 & 74.2 & 97.7 & 73.3 & 98.2 & 85.9 \\
    Gaussian & \textbf{86.3} & 87.5 & \textbf{75.0} & \textbf{98.0} & \textbf{74.6} & \textbf{98.9} & \textbf{86.7} \\
    \bottomrule
  \end{tabular}
\end{table}

\textbf{Feature Visualizations and Target Discriminability.}
To further evaluate the quality of learned representations, we visualize the representations of KECA model and Oracle model (i.e., ground-truth weight $w^*$). Besides, we also use the values of linear discriminant analysis (LDA)~\cite{hastie2009elements} to evaluate the discriminability on the target domain which implies the generalization of source discriminative knowledge. Specifically, the inter-class scatter, intra-class scatter and LDA value (i.e., inter-class scatter divided by intra-class scatter) on the target domain are reported. The results are shown in Fig.~\ref{fig:feavis_31_AD_KECA}-\ref{fig:feavis_31_AW_oracle}. It can be observed that the cross-domain representations of both KECA and Oracle are appropriately aligned, and the clusters structures are matched. Besides, the average discriminability values of KECA and Oracle are nearly the same, which implies that the proposed method effectively alleviates the biased risk estimation induced by label shift. These results demonstrate that the KECA model with estimated weight indeed serves as a good approximation for the Oracle model with ideal importance weight $w^*$, and ensures the discriminability of the learned target representations.

\textbf{Label/Conditional Discrepancy Optimization.}
Since Thm.~\ref{thm:GLS-upper-bound} decomposes the sufficient condition of successful transfer as label shift and conditional shift terms, it is important to evaluate the discrepancy values of these shift terms. Though the precision and convergence of BBSE estimator \cite{lipton2018detecting} and CMMD metric \cite{ren2016conditional,song2009hilbert} are theoretically proved under some assumption, e.g., conditional invariant property and linear independence, it is still unclear the influence of estimation error in empirical scenarios. To understand the label discrepancy and conditional discrepancy empirically, we conduct experiments on Office-31 to show the curves of divergence values on label distributions and conditional distributions. From the results on Fig.~\ref{fig:DivCurve_31_AD}-\ref{fig:DivCurve_31_DA}, it can be observed that the label discrepancy decreases rapidly in the initial stage, and the conditional discrepancy is significantly mitigated by the kernel-based alignment. Besides, note that we also introduce a warm-up training strategy to reduce the uncertainty of pseudo labels, where only source empirical risk is minimized. After the warm-up stage, the complete GLS correction objective will be optimized. Since the warm-up epochs are empirically set as 40, we can observe from Fig.~\ref{fig:DivCurve_31_AD_AfterWarmup}-\ref{fig:DivCurve_31_DA_AfterWarmup}, the label discrepancy continues to decrease with the application of conditional matching. It demonstrates that the estimations of label distribution and importance weight can also be benefited from the conditional matching. In conclusion, the proposed kernel-based model is effective in minimizing the shift terms in generalization upper bound and ensuring low generalization error.

\begin{table*}
  \centering
  \caption{Classification accuracies (\%) and TV distances ($d_{\TV}\in[0,1]$) between label distributions on submsampled Office-Home (ResNet-50).}
  \label{tab:comparison_subsample_Home}
  \renewcommand{\tabcolsep}{0.16pc} 
  \begin{tabular}{l|ccccccccccccc}
    \toprule
    \rowcolor{gray!25}
    \textbf{Tasks} & sAr$\to$Cl & sAr$\to$Pr & sAr$\to$Rw & sCl$\to$Ar & sCl$\to$Pr & sCl$\to$Rw & sPr$\to$Ar & sPr$\to$Cl & sPr$\to$Rw & sRw$\to$Ar & sRw$\to$Cl & sRw$\to$Pr & Avg. \\
    \rowcolor{gray!25}
    $d_{\TV}(P_Y,Q_Y)$ & 0.27 & 0.30 & 0.26 & 0.40 & 0.32 & 0.33 & 0.38 & 0.29 & 0.30 & 0.35 & 0.26 & 0.28 & 0.31 \\
    Source~\cite{he2016deep} & 35.7 & 54.7 & 62.6 & 43.7 & 52.5 & 56.6 & 44.3 & 33.1 & 65.2 & 57.1 & 40.5 & 70.0 & 51.3   \\
    DANN~\cite{ganin2016domain} & 36.1 & 54.2 & 61.7 & 44.3 & 52.6 & 56.4 & 44.6 & 37.1 & 65.2 & 56.7 & 43.2 & 69.9 & 51.8 \\
    IWDANN~\cite{combes2020domain} & 39.8 & 63.0 & 68.7 & 47.4 & 61.1 & 60.4 & 50.4 & 41.6 & 72.5 & 61.0 & 49.4 & 76.1 & 57.6 \\
    JAN~\cite{long2017deep} & 34.5 & 56.9 & 64.5 & 46.2 & 56.8 & 59.1 & 50.6 & 37.2 & 70.0 & 58.7 & 40.6 & 72.0 & 53.9 \\
    IWJAN~\cite{combes2020domain} & 36.2 & 61.0 & 66.3 & 48.7 & 59.9 & 61.9 & 52.9 & 37.7 & 70.9 & 60.3 & 41.5 & 73.3 & 55.9 \\
    CDAN~\cite{long2018conditional} & 38.9 & 56.8 & 64.8 & 48.0 & 60.1 & 61.2 & 49.7 & 41.4 & 70.2 & 62.4 & 47.0 & 74.7 & 56.3 \\
    IWCDAN~\cite{combes2020domain} & 43.0 & 65.0 & 71.3 & 52.9 & 64.7 & 66.5 & 54.9 & 44.8 & 75.9 & \textbf{67.0} & 50.5 & 78.6 & 61.2 \\
    MARSg~\cite{rakotomamonjy2021optimal} & 47.2 & 62.2 & 68.7 & 46.0 & 59.8 & 62.7 & 47.8 & 42.4 & 70.0 & 54.0 & 45.8 & 72.6 & 56.6 \\
    OSTAR~\cite{kirchmeyer2022mapping} & 48.4 & 69.5 & 73.3 & 46.5 & 63.4 & 63.1
    & 50.8 & 44.2 & 74.0 & 56.3 & 49.1 & 75.6 & 59.5 \\
    \textbf{KECA} & \textbf{51.0} & \textbf{72.9} & \textbf{77.2} & \textbf{58.0} & \textbf{72.2} & \textbf{73.1} & \textbf{60.5} & \textbf{48.8} & \textbf{79.6} & 63.7 & \textbf{52.5} & \textbf{81.1} & \textbf{65.9} \\
    \bottomrule
  \end{tabular}
\end{table*}

\begin{figure*}[t]
  \centering
  \subfloat[Convergence sA$\to$D \label{fig:ConvergeCurve_31_AD}]{\includegraphics[width=0.27\linewidth,trim=75 40 50 70,clip]{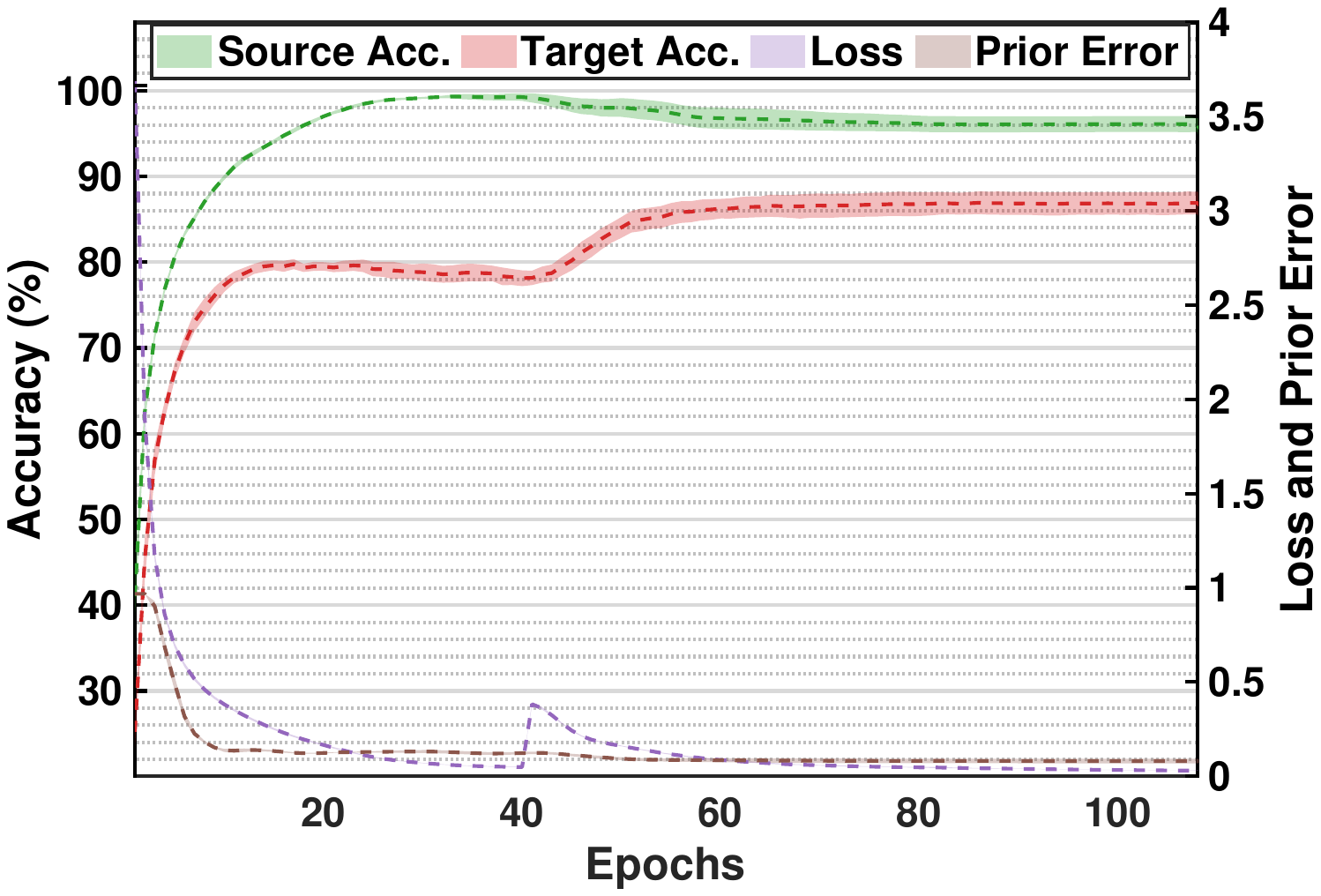}}
  \hfill
  \subfloat[After warm-up sA$\to$D \label{fig:ConvergeCurve_31_AD_AfterWarmup}]{\includegraphics[width=0.27\linewidth,trim=75 40 50 70,clip]{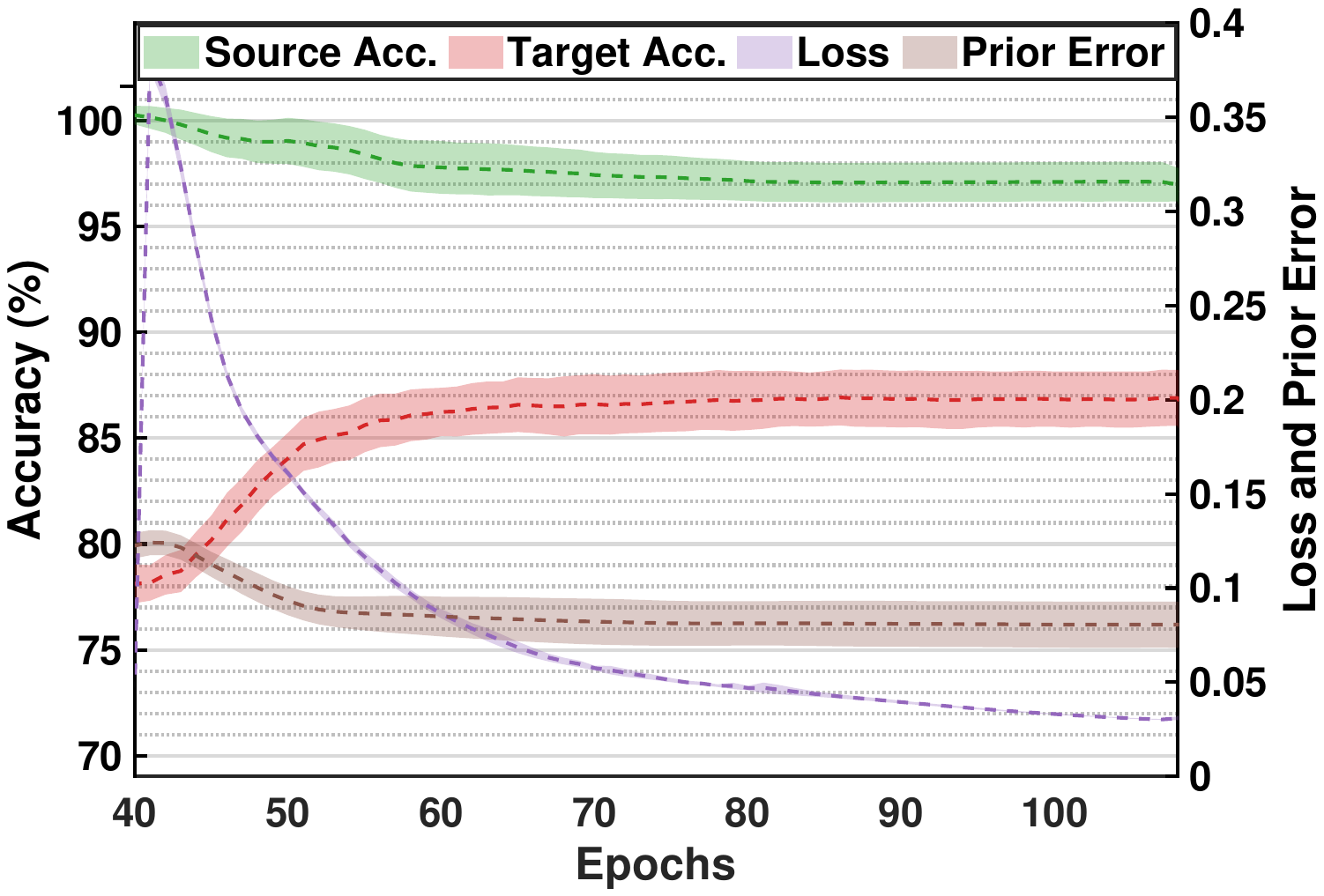}}
  \hfill
  \subfloat[Comparison sA$\to$D \label{fig:ConvergeCurve_31_AD_Compare}]{\includegraphics[width=0.27\linewidth,trim=75 40 80 70,clip]{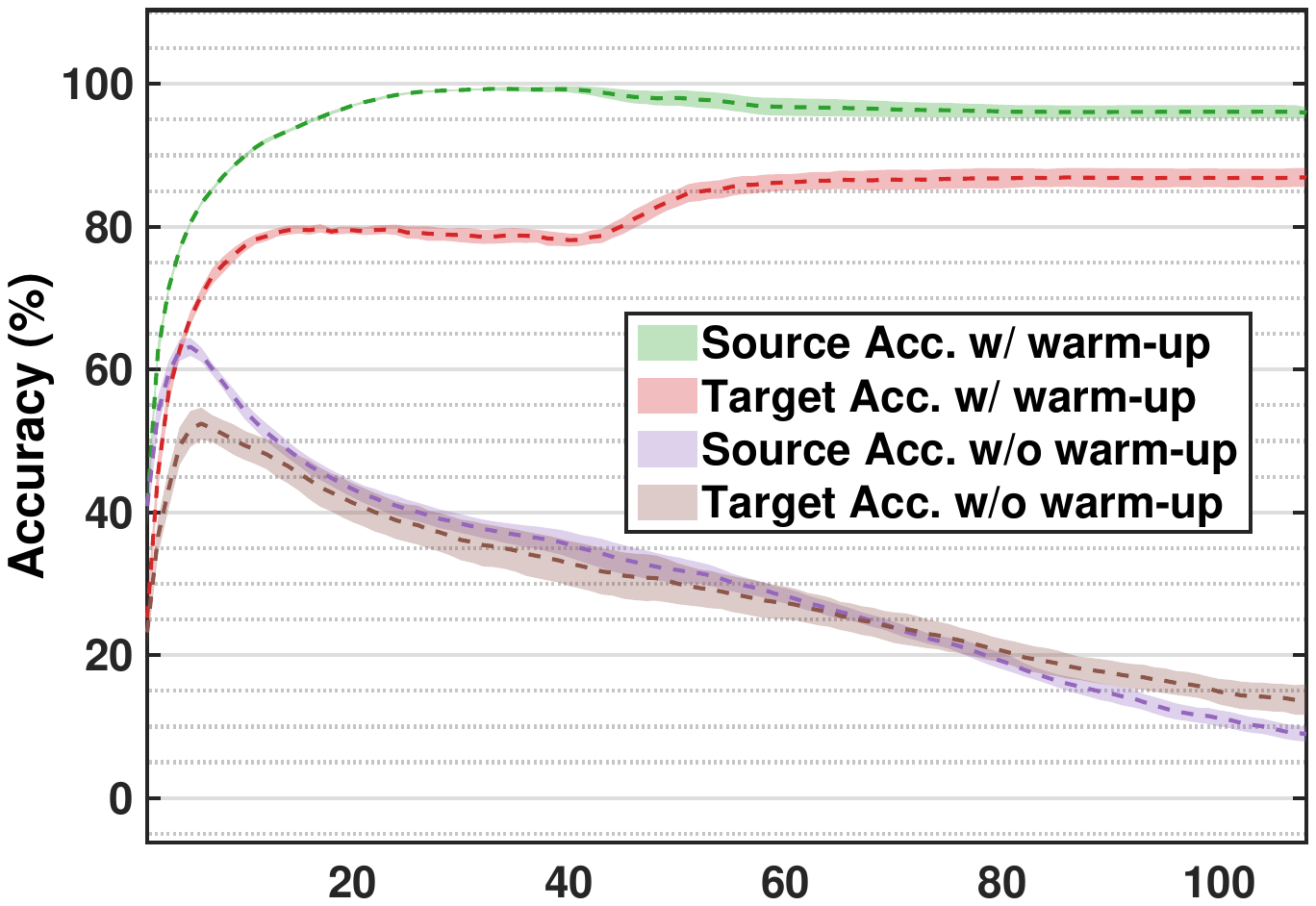}}
  \\
  \centering
  \subfloat[Convergence sD$\to$A \label{fig:ConvergeCurve_31_DA}]{\includegraphics[width=0.27\linewidth,trim=75 40 50 70,clip]{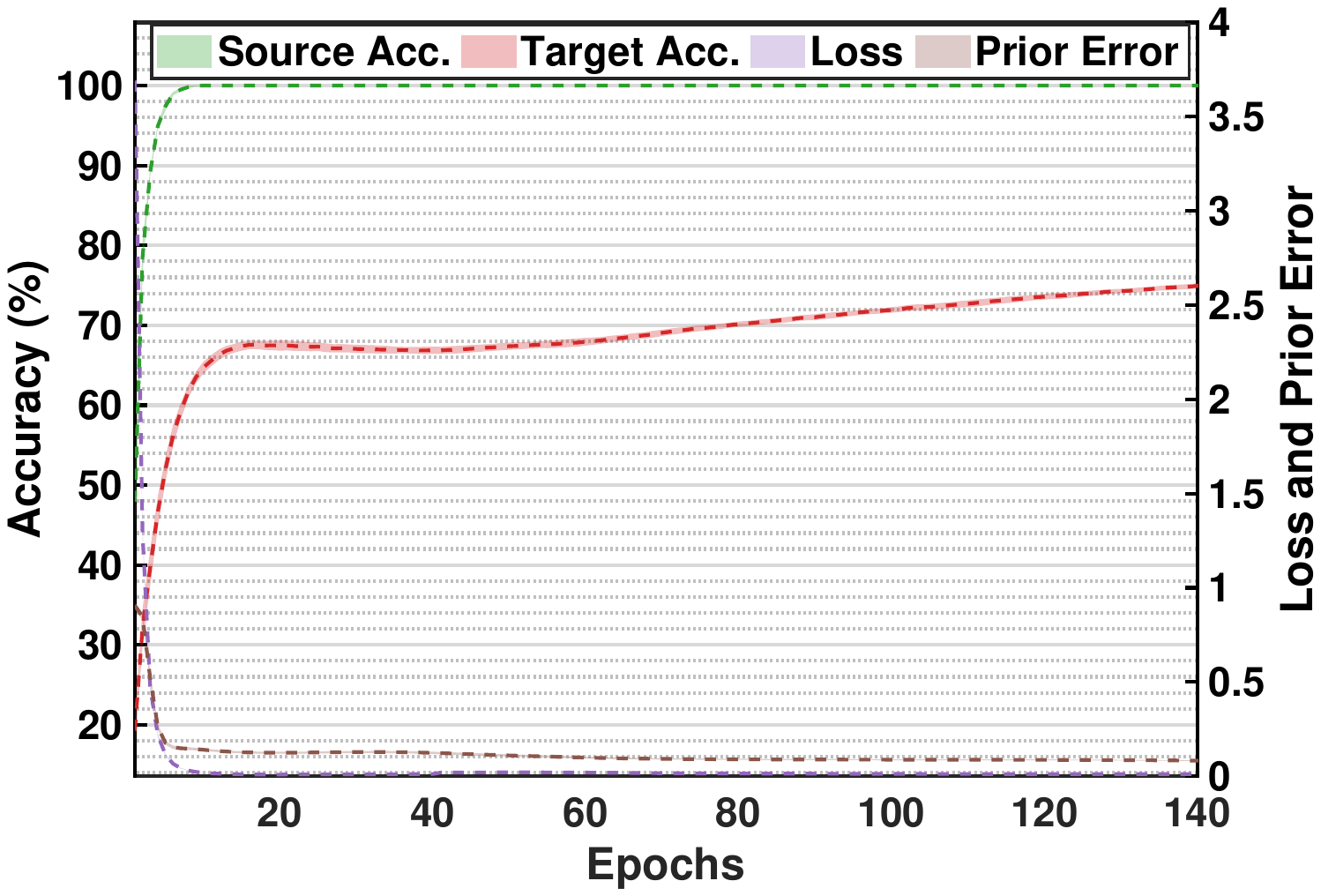}}
  \hfill
  \subfloat[After warm-up sD$\to$A \label{fig:ConvergeCurve_31_DA_AfterWarmup}]{\includegraphics[width=0.27\linewidth,trim=75 40 50 70,clip]{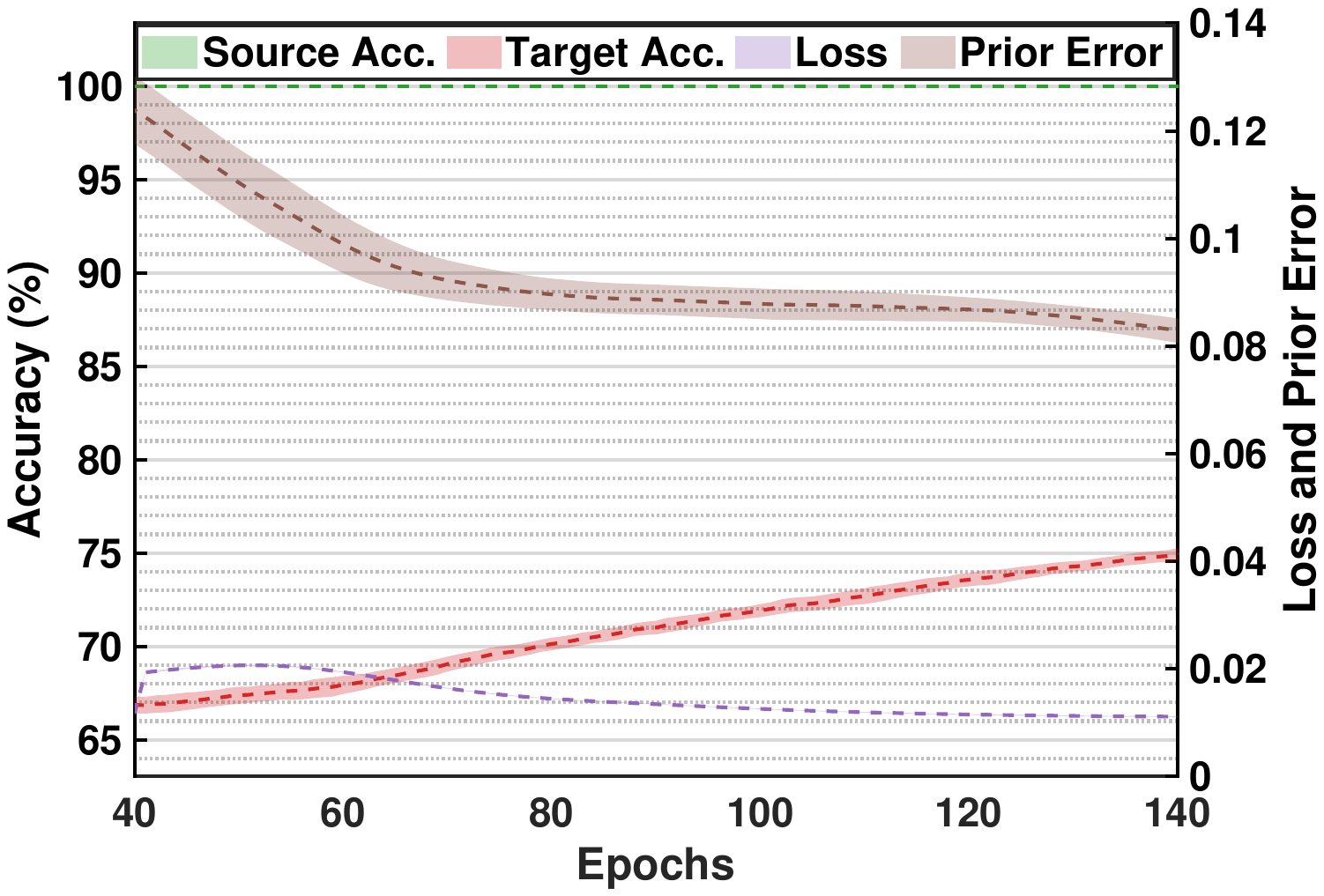}}
  \hfill
  \subfloat[Comparison sD$\to$A \label{fig:ConvergeCurve_31_DA_Compare}]{\includegraphics[width=0.27\linewidth,trim=75 40 80 70,clip]{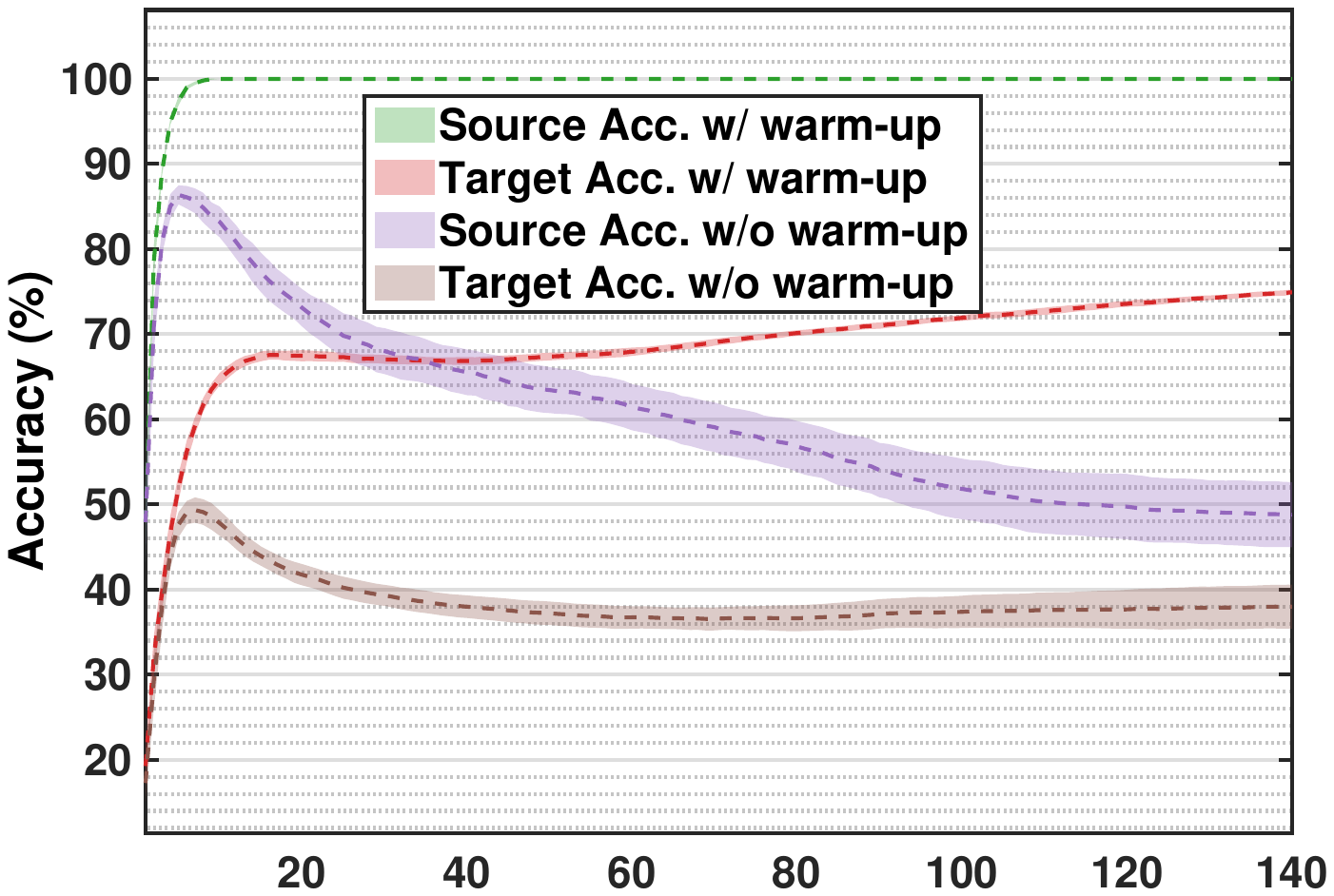}}
  \\
  \caption{Curves of source accuracy, target accuracy, loss objective $\MC{L}_{\mathrm{GLS}}$ and error of prior estimation $d_{\TV}(P^w_Y,Q_Y)$ on Office-31 dataset. (a) and (d): curves of complete training process; (b) and (e): curves after the warm-up stage where the conditional matching and importance weight are applied; (c) and (f): comparison between model w/ warm-up and model w/o warm-up.}
  \label{fig:Converge_Curve}
\end{figure*}

\textbf{Discrepancy Assumption.}
In Thm.~\ref{thm:GLS-lower-bound}, an assumption on label discrepancies is made in the proof of necessity, so it is necessary to observe the assumption in empirical scenario. Though the label discrepancy is easy to compute, the divergence between representations is hard to estimate since the inference for high-dimensional continuous probability distributions is a computationally expensive problem. Fortunately, our theoretical result (i.e., Lem.~\ref{lem:assumption_on_JSD}) shows that the divergence assumption in Thm.~\ref{thm:GLS-lower-bound} holds for any $w$ if the transformation is conditional invariant. It implies that we can alternatively observe the conditional discrepancy which is sufficient to ensure the assumption in Thm.~\ref{thm:GLS-lower-bound}. Therefore, we also show the ground-truth conditional discrepancy $D(P^w_{Z|Y},Q_{Z|Y})$ based on CMMD. The results in Fig.~\ref{fig:DivCurve_31_AD}-\ref{fig:DivCurve_31_DA} show that the conditional discrepancies is effectively minimized by the kernel-based alignment, which implies Lem.~\ref{lem:assumption_on_JSD} can be ensured by condition shift correction. Thus, the label discrepancy assumption in Thm.~\ref{thm:GLS-lower-bound} is generally reliable in empirical scenarios.

\begin{table*}
  \centering
  \caption{Classification accuracies (\%) and TV distances between label distributions on submsampled Office-31 and VisDA-2017 (ResNet-50).}
  \label{tab:comparison_subsample_31_VisDA}
  \renewcommand{\tabcolsep}{0.8pc} 
  \begin{tabular}{l|ccccccc|ccc}
    \toprule
    \rowcolor{gray!25}
    \textbf{Datasets} & \multicolumn{7}{>{\columncolor[gray]{.88}}c|}{\textbf{Office-31}}
    & \multicolumn{3}{>{\columncolor[gray]{.88}}c}{\textbf{VisDA-2017}}\\
    \rowcolor{gray!25}
    \textbf{Tasks} & sA$\to$D & sA$\to$W & sD$\to$A & sD$\to$W & sW$\to$A & sW$\to$D & Avg. & S$\to$R & sS$\to$R & Avg. \\
    \rowcolor{gray!25}
    $d_{\TV}(P_Y,Q_Y)$ & 0.29 & 0.26 & 0.28 & 0.29 & 0.28 & 0.30 & 0.28 & 0.19 & 0.31 & 0.25 \\
    Source~\cite{he2016deep} & 75.8 & 70.7 & 56.8 & 95.3 & 58.4 & 97.3 & 75.7 & 48.4 & 49.0 & 48.7 \\
    DANN~\cite{ganin2016domain} & 75.5 & 77.7 & 56.6 & 93.8 & 57.5 & 96.0 & 76.2 & 61.9 & 52.9 & 57.4 \\
    IWDANN~\cite{combes2020domain} & 81.6 & \textbf{88.4} & 65.0 & 97.0 & 64.9 & 98.7 & 82.6 & 63.5 & 60.2 & 61.9 \\
    JAN~\cite{long2017deep} & 77.7 & 77.6 & 64.5 & 91.7 & 65.1 & 92.6 & 78.2 & 57.0 & 50.6 & 53.8 \\
    IWJAN~\cite{combes2020domain} & 84.6 & 83.3 & 65.3 & 96.3 & 67.4 & 98.8 & 82.6 & 57.6 & 57.1 & 57.3 \\
    CDAN~\cite{long2018conditional} & 82.5 & 84.6 & 62.5 & 96.8 & 65.0 & 98.3 & 81.6 & 65.6 & 60.2 & 62.9 \\
    IWCDAN~\cite{combes2020domain} & \textbf{86.6} & 87.3 & 66.5 & 97.7 & 66.3 & \textbf{98.9} & 83.9 & 66.5 & 65.8 & 66.2 \\
    MARSg~\cite{rakotomamonjy2021optimal} & 84.5  & 81.6  & 62.1  & 91.0  & 63.9  & 98.0  & 80.2 & 55.6 & 55.0 & 55.3 \\
    OSTAR~\cite{kirchmeyer2022mapping} & 84.2  & 83.9  & 65.0  & 94.1  & 70.0  & 98.5  & 82.6 & 59.2 & 58.8 & 59.0 \\
    \textbf{KECA} & 86.3 & 87.5 & \textbf{75.0} & \textbf{98.1} & \textbf{74.8} & \textbf{98.9} & \textbf{86.8} & \textbf{73.3} & \textbf{71.4} & \textbf{72.4}  \\
    \bottomrule
  \end{tabular}
\end{table*}

\textbf{Pseudo Label Analysis.}
In empirical scenarios, since the target domain is unlabeled, the pseudo labels are usually necessary to explore label information on target domain and learn conditional invariant property approximately \cite{long2018conditional,combes2020domain}. But, the pseudo labels are usually error-prone in the initial stage, which leads to the unreliable estimations of conditional discrepancy and importance weights. To overcome this problem, we develop a warm-up stage with few epochs as discussed in Sec.~\ref{subsec:set_up}, where the model is only trained on the source domain with classification loss. To empirically observe the uncertainty in training process and the reliability of pseudo labels, we conduct experiment to analysis the convergence of KECA model by visualizing the training curves of source accuracy, target accuracy, loss objective $\MC{L}_{\mathrm{GLS}}$ and error of prior estimation $d_{\TV}(P^w_Y,Q_Y)$. 

We first show that the pseudo labels are indeed practical for empirical application. From the results in Fig.~\ref{fig:ConvergeCurve_31_AD}-\ref{fig:ConvergeCurve_31_AD_AfterWarmup} and~\ref{fig:ConvergeCurve_31_DA}-\ref{fig:ConvergeCurve_31_DA_AfterWarmup}, it can be observed that the accuracies on the target domain are significantly improved ($\sim$10\%) after the warm-up stage, i.e., 40 epochs, which demonstrates that the pseudo labels can provide reliable label information for learning conditional invariant model; it also validates that GLS correction with pseudo labels is sufficient to ensure better generalization performance on the target domain. Besides, the label estimation errors consistently decrease after the warm-up stage, which validates that the learned predictor also ensures better label shift correction.

Further, we can conclude that the warm-up strategy is effective in improving the quality of pseudo labels, which ensures appropriate shift correction. Specifically, as shown in the comparison in Fig.~\ref{fig:ConvergeCurve_31_AD_Compare}~and~\ref{fig:ConvergeCurve_31_DA_Compare}, if the conditional alignment and importance weighting strategy are applied in the initial stage with randomly initialized parameters, i.e., model w/o warm-up, the performance is significantly lower than the model with warm-up. Besides, the accuracies on the target domain continue to increase after the warm-up stage, while the model without warm-up usually suffers from the performance degradation. Therefore, the warm-up training is indeed necessary to mitigate the risk in pseudo labels.  

In conclusion, these results demonstrate that the convergences of label estimation error and model's accuracy are generally ensured in empirical scenarios, and the GLS correction with pseudo labels are effective.

\textbf{Performance Comparison.}
To compare our model with other SOTA methods, we present the results of two types of dataset shift methods in Tab.~\ref{tab:comparison_subsample_Home}-\ref{tab:comparison_subsample_31_VisDA}. 1) Invariant representation learning: DANN~\cite{ganin2016domain}, JAN~\cite{long2017deep}, CDAN~\cite{long2018conditional}; 2) GLS correction: Importance Weighted (IW) invariant representation learning~\cite{combes2020domain}, MARSg~\cite{rakotomamonjy2021optimal} and OSTAR~\cite{kirchmeyer2022mapping}. The TV distances between $P_Y$ and $Q_Y$ are also provided to quantify the degrees of label shift on different transfer tasks. 

From the results, we can observe that the invariant representation learning models (i.e., DANN, JAN and CDAN) improve the accuracies of standard learning model (i.e., Source) by about 0.5\%$\sim$14\% in average, where the conditional model CDAN is generally better compared with marginal model DANN. By applying weight $w$, the IW variants further achieve higher accuracies compared with original invariant representation learning models, and improve the mean accuracies on Office-Home, Office-31 and VisDA-2017 to 61.2\%, 83.9\% and 66.2\%, respectively. Besides, the OT-based GLS models MARSg and OSTAR also ensure better performance than invariant representation learning models, and improve the mean accuracies to 59.5\%, 82.6\% and 59.0\%. These results validate that invariant transformation is insufficient for dataset shift correction (i.e., Prop.~\ref{prop:imposs-joint-matching}) and $g_{\mar}$ is generally error-prone (i.e., Rem.~\ref{rem:imposs_marginal}). For KECA, it explicitly alleviates the GLS and achieving the highest accuracies on all datasets and most of the transfer tasks. Compared with other SOTA GLS correction models, KECA improves the mean accuracies by 4\%$\sim$6\%, and achieves the mean accuracies of 65.9\%, 86.8\% and 72.4\%. The reason is that KECA directly minimizes the generalization upper bound which ensures the sufficiency for dealing with dataset shift as Thm.~\ref{thm:GLS-upper-bound}. Besides, note that VisDA-2017 contains a larger visual shift compared with Office-31 and Office-Home datasets. Then the significant performance improvement on VisDA-2017 demonstrates that KECA can ensure a better transformation to learn the conditional invariant property and preserve the semantic information under large dataset shift. Overall, the comparison experiments verify the validity of our theoretical results and learning algorithm on GLS correction.

\section{Conclusions}
\label{sec:conclusions}
In this paper, we systematically study the dilemma of invariant representation learning under label shift and present a theoretical solution with GLS correction. The main results suggest that learning transformation is error-prone under label shift and a GLS correction triplet is sufficient and necessary for successful knowledge transfer. These results provide insights into learning transferable model under general scenarios. Besides, a unified shift correction framework and a theory-driven algorithm are proposed to minimize the derived generalization upper bound and learn a sufficient model for transfer. The validity of theory and algorithm is verified by numerical experiments and analysis. For future directions, developing GLS correction algorithms for the scenarios that target data are unavailable, e.g., domain generalization or source-free scenarios, will be a crucial step towards real-world applications.


\bibliography{GLS_Theory_Ref}
\bibliographystyle{IEEEtr}

\begin{IEEEbiography}[{\resizebox{1.0in}{1.3in}{\includegraphics*{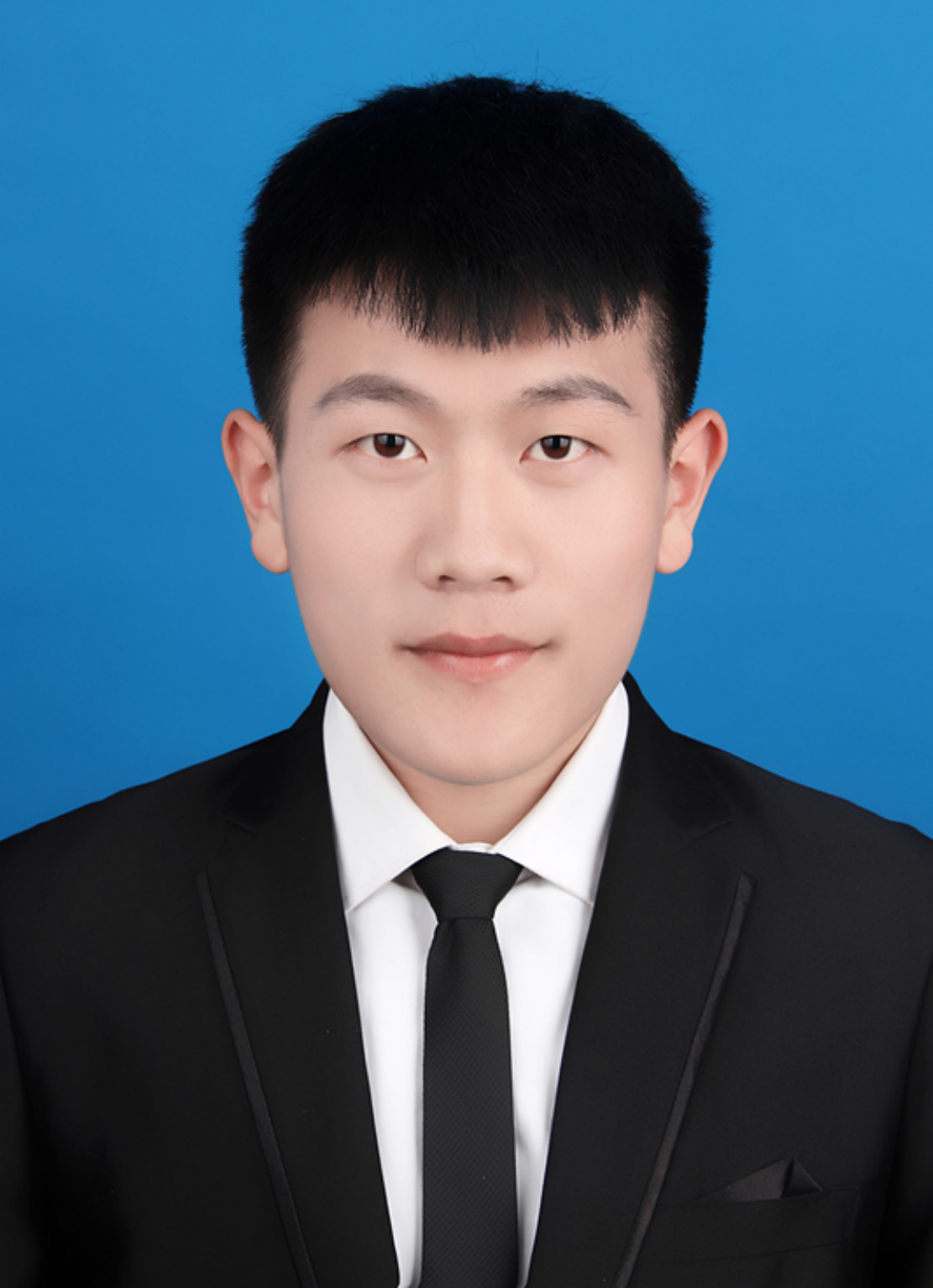}}}]{You-Wei Luo}
received the B.S. degree in Statistics from China University of Mining and Technology, Xuzhou, China, in 2018 and the Ph.D. degree in Mathematics from Sun Yat-sen University, Guangzhou, China, in 2023. \par He is currently a post-doctoral fellow with the School of Mathematics, Sun Yat-sen University, Guangzhou, China. His research interests include image processing, manifold learning and transfer learning.
\end{IEEEbiography}


\begin{IEEEbiography}[{\resizebox{1.0in}{1.3in}{\includegraphics*{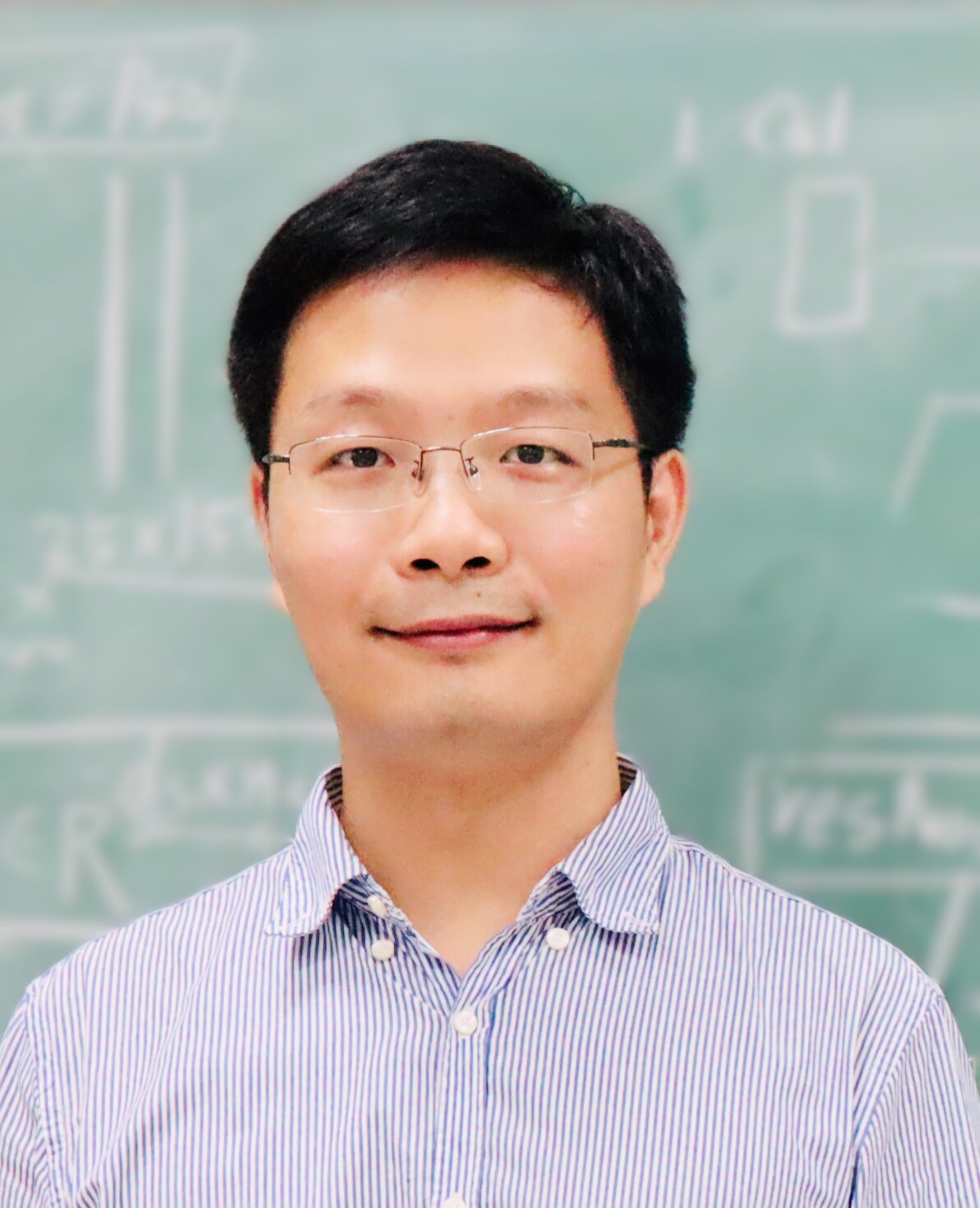}}}]{Chuan-Xian Ren}
received the Ph.D. degree in Mathematics from Sun Yat-Sen University, Guangzhou, China, in 2010. During 2010 and 2011, he was with the Department of Electronic Engineering, City University of Hong Kong, as a Senior Research Associate. \par He is currently a Professor of the School of Mathematics, Sun Yat-Sen University. His research interests include visual pattern analysis and machine learning.
\end{IEEEbiography}

\end{document}